\documentclass[conference]{IEEEtran}
\usepackage[sort, numbers]{natbib}
\usepackage{booktabs}
\usepackage{graphicx}
\usepackage{amsmath}
\usepackage{amssymb}
\usepackage{subcaption}
\usepackage[font=footnotesize]{caption} %
\usepackage[show]{notes} %
\usepackage{hyperref}
\usepackage{xstring}
\usepackage[normalem]{ulem}
\usepackage{array}
\usepackage{url}
\usepackage[utf8]{inputenc}
\usepackage{mathtools,xparse}
\usepackage{graphicx,dblfloatfix}     %

\newcounter{tisctr}
\renewcommand{\thetisctr}{\arabic{tisctr}}

\usepackage{array}
\newcolumntype{H}{>{\setbox0=\hbox\bgroup}c<{\egroup}@{}}

\DeclarePairedDelimiterX{\norm}[1]{\lvert}{\rvert}{#1}
\DeclarePairedDelimiterX{\normF}[1]{\lVert}{\rVert}{#1}

\newcommand{\spara}[1]{\smallskip\noindent\textbf{#1}}

	{\normalsize \end{list} 
		\vskip 1pt}

\newenvironment{defn-eqn}[2]{\vskip 6pt \refstepcounter{tisctr} %
	\noindent {\textbf{Definition \thetisctr.~}(#1) \emph{#2}} \vspace{2mm}}
	
\newenvironment{defn-test}[2]{\vskip 6pt \refstepcounter{tisctr} %
	\noindent {{\bf Definition \thetisctr.~}(#1) \emph{#2}}%
	\vskip 0pt
	\begin{equation}{}{\labelwidth 0pt \labelsep 0pt%
			\parsep 0pt
		}}%
		{\normalsize \end{equation} 
	    \vspace{-1mm}%
		}	
		
\newcommand{\squishlist}{
	\begin{list}{$\bullet$}
		{ \setlength{\itemsep}{0pt}
			\setlength{\parsep}{1pt}
			\setlength{\topsep}{1pt}
			\setlength{\partopsep}{0pt}
			\setlength{\leftmargin}{1.0em}
			\setlength{\labelwidth}{1em}
			\setlength{\labelsep}{0.5em} } }
	
	\newcommand{\squishend}{\end{list}}

\usepackage{color}
\usepackage{balance}
\usepackage{comment}
\usepackage{wrapfig}

\def\BibTeX{{\rm B\kern-.05em{\sc i\kern-.025em b}\kern-.08em
		T\kern-.1667em\lower.7ex\hbox{E}\kern-.125emX}}

\begin{document}

\title{
	\vspace{-5mm}
	Detecting and Mitigating Test-time Failure Risks
	via Model-agnostic Uncertainty Learning}

\author{
	\IEEEauthorblockN{Preethi Lahoti}
	\IEEEauthorblockA{Max Planck Institute for Informatics \\
		plahoti@mpi-inf.mpg.de}	
	\vspace{-7mm}
	\and
	\IEEEauthorblockN{Krishna P. Gummadi}
	\IEEEauthorblockA{Max Planck Institute for Software Systems\\
		gummadi@mpi-sws.org}
	\vspace{-7mm}
	\and
	\IEEEauthorblockN{Gerhard Weikum}
	\IEEEauthorblockA{Max Planck Institute for Informatics\\
		weikum@mpi-inf.mpg.de}
	\vspace{-7mm}
}

\maketitle

\begin{abstract}
Reliably predicting potential failure risks of machine learning (ML)
systems when deployed with production data
is a crucial aspect of trustworthy AI. 
This paper introduces \emph{Risk Advisor},
a novel post-hoc
\emph{meta-learner}
for estimating failure risks and predictive uncertainties of {\em any already-trained} black-box classification model.
In addition to providing a {\em risk score}, the {\em Risk Advisor} decomposes the uncertainty estimates into aleatoric and epistemic uncertainty components, thus giving 
informative insights into the sources of uncertainty inducing the failures. 
Consequently, {\em Risk Advisor} can distinguish between failures caused by data variability, data shifts and model limitations 
and advise on mitigation actions (e.g., collecting more data to counter data shift).
Extensive experiments on various families of black-box classification models 
and on 
real-world and synthetic datasets covering common ML failure scenarios
show that the
\emph{Risk Advisor}
reliably
predicts deployment-time failure risks
in all the scenarios,
and outperforms
strong
baselines.
\end{abstract}

\maketitle
\section{Introduction} %

\noindent {\bf Motivation and Problem:} 
Machine learning (ML) systems have found wide adoption in mission-critical applications. 
Their success crucially hinges on the amount and quality of training data, and also on the
assumption that the data distribution for the deployed system stays the same and is well
covered by the training samples.
However, this cannot be taken for granted. 
\citet{saria2019tutorial} categorize limitations and failures of ML systems into several regimes,
including data shifts (between training-time and deployment-time distributions),
high data variability (such as overlapping class labels near decision boundaries)
and model limitations (such as log-linear decision boundaries vs. neural ML).
Trustworthy ML needs models and tools for detecting such failure risks and analyzing the
underlying sources of uncertainty.
Unfortunately,
systems often fail silently without any warning, 
despite
showing high confidence in their predictions \cite{nguyen2015deep, jiang2018trust, goodfellow2014explaining}.

This paper addresses the challenge of predicting, analyzing and mitigating 
failure risks for classifier systems.
The goal is to provide the system with \emph{uncertainty scores} for its predictions, 
so as to (a) reliably predict test-time inputs for which the system is likely to fail, and (b) 
detect the \emph{type of uncertainty} that induces the risk, so that (c) appropriate 
\emph{mitigation actions} can be pursued.
Equipped with different kinds of {uncertainty scores},
a deployed system could 
improve its robustness
in handling new data points that pose difficult situations.
For instance, if an introspection component
indicates that the ML system's output has a non-negligible likelihood
of being erroneous, the system could abstain and defer the decision to a human expert
(rather than risking adverse effect on human lives).
When many production data points are out-of-distribution,
collecting additional training samples and retraining the system
would be a remedy.
The challenge here is to determine which action is advised under which conditions.
This is the problem addressed in this paper:
determine the amount and type of uncertainty in deployment-time inputs, so as to decide
if and which kind of mitigation is needed.

\begin{figure}[tbh!]
	\begin{subfigure}{0.22\columnwidth}
		\centering	
		\includegraphics[scale=.6]{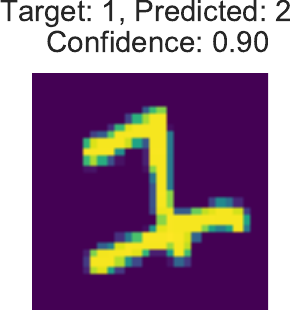}
		\caption{}
		\vspace{-1mm}%
		\label{fig:fashion-mnist-aleatoric-a}
	\end{subfigure}	
	\begin{subfigure}{0.22\columnwidth}
		\centering	
		\includegraphics[scale=.6]{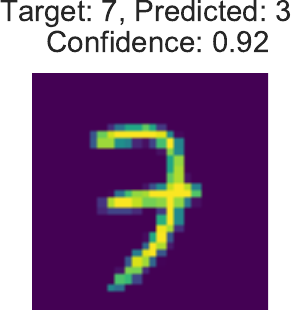}
		\caption{}
		\vspace{-1mm}%
		\label{fig:fashion-mnist-aleatoric-b}
	\end{subfigure}
	\begin{subfigure}{0.26\columnwidth}
		\centering	
		\includegraphics[scale=.59]{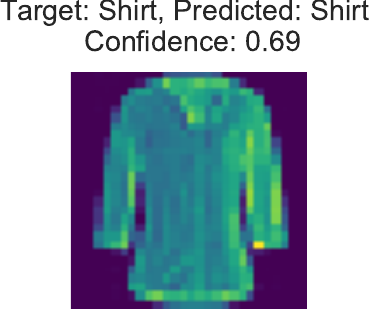}
		\caption{}
		\vspace{-1mm}%
		\label{fig:fashion-mnist-shirt-a}
	\end{subfigure}	
	\begin{subfigure}{0.26\columnwidth}
		\centering	
		\includegraphics[scale=.59]{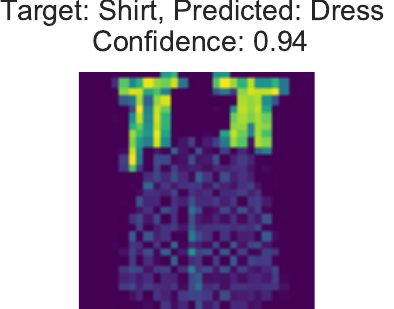}
		\caption{}
		\vspace{-1mm}%
		\label{fig:fashion-mnist-shirt-b}
	\end{subfigure}	
	\caption{Examples of ground-truth (target) and predicted labels where 
(a,b) 
a CNN fails despite high confidence (MNIST dataset \cite{lecun2010mnist}),
and  (c,d) a CNN assigns higher confidence to a misclassified sample than to a correct one
(Fashion MNIST dataset \cite{fashion-mnist})}
	\label{fig:mnist-examples}
	\vspace{-3mm}%
\end{figure}

\vspace*{0.1cm}
\noindent {\bf State of the Art and its Limitations:}
The standard approach for deciding whether an ML system's predictions are trustworthy is based on
confidence scores computed over predictive probabilities, such as max class probability (MCP) in neural ML \cite{hendrycks2016baseline}
or distance from the decision boundary for SVM.
However, predictive probabilities are 
not reliable
estimates of a model's uncertainty \cite{gal2016dropout, jiang2018trust,nguyen2015deep, goodfellow2014explaining}.
Figure \ref{fig:mnist-examples}(a,b) shows two examples where a CNN model misclassifies handwritten digits (from the MNIST benchmark) while
giving high scores for its (self-) confidence.
Even if the confidence scores are calibrated (e.g., \cite{platt1999probabilistic,guo2017calibration}), 
they may still not be trustworthy as the ordering of the confidence scores can itself be unreliable.
This is because calibration methods are concerned with scaling of the scores,
i.e., they perform monotonic transformations with respect to prediction scores, which do not alter the ranking of confident vs. uncertain example.
Fig. \ref{fig:mnist-examples}(c,d) shows two examples where a CNN model gives higher score to a
misclassified sample than to a correct one. 

More importantly, {confidence scores} do not reflect what the model {\em does not know}.
In Fig. \ref{fig:mnist-examples}(c,d), the Fashion MNIST dataset has many positive training examples of shirts similar to (c) while hardly any examples that resemble (d) -- a case where the training distribution does not sufficiently reflect the test-time data.
Yet, the CNN model makes a prediction with high confidence of 0.94 (see Fig. \ref{fig:mnist-examples}d).
This limitation holds even for the state-of-the-art model {\em Trust Score} \cite{jiang2018trust}, which serves as a 
major baseline for this paper.

Moreover and most critically,
{\em confidence scores} are ``one-dimensional''
and do not provide insight on which type of uncertainty
is the problematic issue. Thus, confidence scores from prior works are limited in their support for identifying different types of 
appropriate mitigation.

A common line of work for uncertainty estimation builds on Bayesian methods \cite{denker1990transforming, barber1998ensemble}, or making specialized changes to the learning algorithm (e.g.,\cite{gal2016dropout, depeweg2018decomposition,lakshminarayanan2017simple,shaker2020aleatoric,ustimenko2020uncertainty}).
However, these 
are tightly coupled to the choice of the
underlying classification model and thus involve making specialized modifications to the ML pipeline.
Therefore, 
such techniques are 
unsuitable for dealing with a broad variety of black-box ML systems.

\begin{figure}[tbh]
	\centering	
	\includegraphics[scale=0.47]{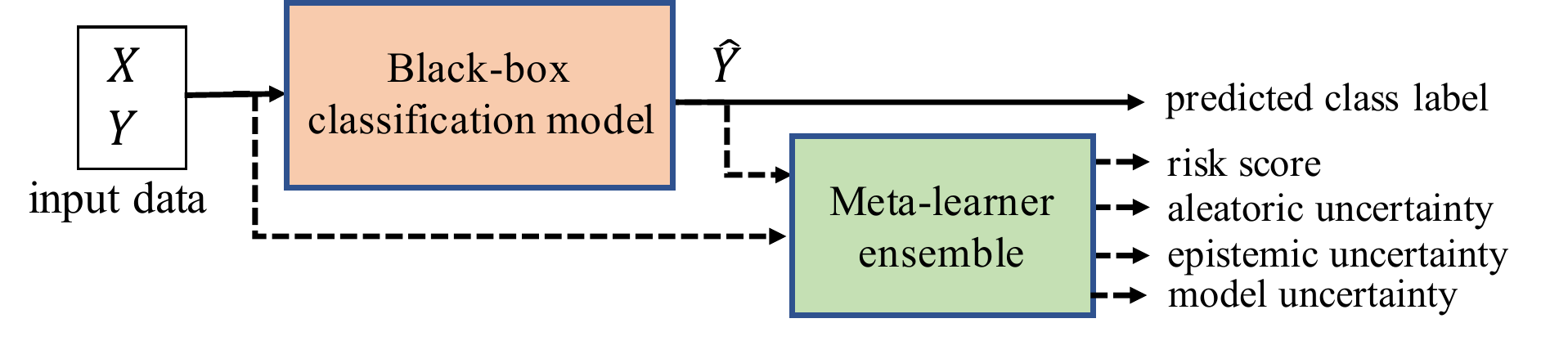}
	\vspace{-3mm}%
	\caption{Schematic overview of the Risk Advisor framework.}
	\label{fig:blockdiagram}
\end{figure}

\vspace*{0.1cm}
\noindent{\bf Proposed Approach:}
This paper presents
\emph{Risk Advisor}, a generic and 
versatile framework for reliably estimating failure risks of any {already-trained} black-box classification model.
The {\em Risk Advisor} consists of a post-hoc {\em meta-learner} for uncertainty estimation that is separate from the underlying ML system, and can be incorporated without any code changes in the underlying
ML pipeline.
The {\em meta-learner} is model-agnostic: it can be applied to any family of black-box classifiers (e.g., deep neural networks, decision-trees, etc).
Fig. \ref{fig:blockdiagram} gives a 
schematic overview of our framework.

In addition to providing a {\em risk score}
that is more reliable than those of prior works,
the {\em Risk Advisor} provides a refined analysis
of the underlying types of uncertainty inducing the risks.
To this end, we make use of the information-theoretic notions
of {\em model uncertainty}, {\em aleatoric uncertainty} and
{\em epistemic uncertainty} \cite{hora1996aleatory, der2009aleatory,senge2014reliable}.
These concepts are fairly old, but to the best of our knowledge,
have not been considered 
for risk analysis of black-box ML systems. 
Our {\em Risk Advisor} quantifies each of the three risk types and thus
enables judicious 
advise on risk mitigation action, depending
on the type of uncertainty inducing the risks:
\squishlist
\item {\em Aleatoric uncertainty} reflects the variability of data points and the resulting noise around the classifier's decision boundary. A high value indicates that it is inherently difficult to distinguish the output classes, and an appropriate
mitigation then is to equip the deployed system with the option to {\em abstain} rather than forcing an output label.
Fig. \ref{fig:mnist-examples}(a,b) is a case of high aleotoric uncertainty.
\item {\em Epistemic uncertainty} captures systematic gaps in the training samples, like regions where training samples are sparse but have a substantial population of test points after deployment.
This situation can only be countered by obtaining {\em more training data} for the underrepresented critical regions. 
Fig. \ref{fig:mnist-examples}(c,d) is a case of high epistemic uncertainty.
\item {\em Model uncertainty} is an indicator that the black-box ML system 
uses 
models with insufficient learning capacity. In this situation, the proper action is to re-build the ML system
with 
higher model capacity or 
a more expressive learning model, for example, a deep neural network instead of a log-linear model or
Transformer instead of LSTM or CNN.
\squishend

The proposed {\em meta-learner} for estimating the different types of uncertainty
in the {\em Risk Advisor} framework
is implemented as an ensemble of $M$ stochastic gradient-boosted decision trees (E-SGBT).
Each stochastic gradient boosted tree (SGBT) operates on the input-output pairs of training samples
and an indicator variable stating whether the 
trained 
black-box ML system
misclassified 
the training point.
The {\em Risk Advisor's} 
analysis of 
uncertainty is based on the ensemble's ability to 
compute aleatoric and epistemic uncertainty.
All of the uncertainty scores are computed on the training data,
and also at deployment time for test data alone to identify 
slowly evolving risks.

\smallskip
\noindent
\textbf{Contributions:} 
The state-of-the-art method to which we compare our approach
is {\em Trust Score} \cite{jiang2018trust}, which also operates in a model-agnostic post-hoc way.
Trust Score is based on the distance between a test point and its nearest neighbors in the training data.
Its output is a single value, which does not provide guidance on identifying the type of risk.

\smallskip
This paper's novel contributions are as follows:
\squishlist
\item We introduce the \emph{Risk Advisor} framework, the first \emph{model-agnostic} 
method
to detect and mitigate deployment-time failure risks,
requiring access  {only}  to the base 
classifier's training data and its predictions and coping with any kind of underlying Black-box ML model.
\item The \emph{Risk Advisor} is the first method that
leverages the information-theoretic notions of \emph{aleatoric} and \emph{epistemic} uncertainty to 
distinguish between 
ML model failures caused by
distribution shifts between training data and deployment data, 
inherent data variability, and model limitations.
\item Experiments with synthetic and real-world datasets show that
our approach
successfully detects 
uncertainty and failure risks
for many families of ML classifiers, including deep neural models, and does so better than prior baselines including the Trust Score method
by \citet{jiang2018trust}.
\item We demonstrate the 
{\em Risk Advisor}'s practical utility
by three kinds of risk mitigation: (i) selectively abstaining from making predictions under uncertainty (ii) detecting out-of-distribution test-examples (iii) countering risks due to data shift by collecting more training samples in a judicious way.
\squishend 
\section{Related Work}\label{sec:related}

The standard approach for predicting failure risks of ML systems
is to rely on the system's native
(self-) \emph{confidence scores}. 
An implicit assumption is that
that most uncertain data points lie near the decision boundary, and 
confidence increases when moving away from the boundary.
While this is reasonable to capture \emph{aleatoric} uncertainty, this kind of
confidence score fails to capture \emph{epistemic} and \emph{model} uncertainty~\cite{gal2016dropout}.
Techniques for confidence calibration \cite{platt1999probabilistic, guo2017calibration}
are concerned with re-scaling the model's prediction scores to produce calibrated probabilities. 
Like all single-dimensional notions of confidence, this is insufficient to distinguish
different types of uncertainty and resulting risks.
In particular, there is no awareness of
{epistemic} uncertainty 
due to 
data shifts \cite{ovadia2019can}.

Bayesian methods are a common approach to capture uncertainty in ML~\cite{denker1990transforming, barber1998ensemble}.
Recently, a number of 
non-Bayesian specialized learning algorithms were proposed to approximate Bayesian methods. 
For instance, 
variational learning~\cite{honkela2004variational, kendall2017uncertainties}, 
drop-out~\cite{gal2016dropout}, and ensembles of deep neural networks~\cite{lakshminarayanan2017simple}.
However, these models 
tend to be computationally expensive 
(by increasing network size and model parameters), and
are not always practically viable.
Moreover, they require changes to the architecture and code of the underlying ML system.

The concepts of aleatoric and epistemic uncertainty are rooted in statistics and information theory~\cite{hora1996aleatory,der2009aleatory} (\cite{hullermeier2021aleatoric} is a recent overview).
\cite{senge2014reliable} has  incorporated these measures into 
a Bayesian classifier with fuzzy preference modeling.
\cite{shaker2020aleatoric} integrated the distinction between aleatoric and epistemic uncertainty
into random-forest classifiers to enhance its robustness.
Both of these works are focused on one specific ML model and do not work outside these design points,
whereas Risk Advisor is model-agnostic and as such universally applicable.
\cite{shaker2020aleatoric} is included in the baselines for our experimental comparisons.

Several post-hoc approaches were proposed for 
confidence scores 
of already trained classifiers.
\citet{schulam2019can} proposed a post-hoc auditor to learn pointwise reliability scores. However, it is not model-agnostic as it relies on using gradients and the Hessian of 
the underlying ML model.
Further, it does not differentiate between different types of uncertainty.
\citet{schelter2020learning}  proposed a \emph{model-agnostic} validation approach to detect \emph{data-related} errors at serving time. However, 
this work
focuses on 
errors arising from
 data-processing issues, such as missing values or 
incorrectly entered values, and relies on programmatic specification of typical data errors.

The closest approach to ours is \emph{Trust Score} \cite{jiang2018trust}, a model-agnostic method that can be applied post-hoc to any ML system. 
Trust Score measures the agreement between a classifier's predictions
and the predictions of a modified nearest-neighbour classifier which accounts for density distribution.
More precisely, the \emph{trust score} for a new test-time data point is defined as the ratio between (a) the distance from the test sample to its nearest $\alpha$-high density set with a \emph{different} class and (b) the distance from the test sample to its nearest $\alpha$-high density set with the \emph{same} class. A crucial limitation of this approach is that it is 
highly sensitive to
the choice of the distance metric for
defining neighborhoods, and can degrade for high-dimensional data.
Also, it does not provide any guidance on the type of uncertainty.

Classification with reject option \cite{bartlett2008classification} and selective %
abstention
\cite{el2010foundations} are 
related problems,
where the model can defer decisions (e.g., to to a human expert) when it has low confidence.
However, these methods still rely on their own \emph{confidence} scores to determine when to abstain, and thus share the limitations and pitfalls of a single-dimensional self-confidence.
Similarly, the problem of detecting data shifts has been widely studied 
e.g., for detecting and countering {covariate and label shift}\cite{schneider2020improving} and for {anomaly detection} \cite{ben2005outlier,steinwart2005classification}.
These methods address data shifts, but they do not consider failure risks arising from \emph{aleatoric} and \emph{model} uncertainty.

\section{Risk Advisor Model}\label{sec:model}

\subsection{\bf \em Basic Concepts}

\spara{Black-box Classifier's Task:} 
We are given a training dataset $\mathcal{D} = \{ (x_i, y_i)\, \cdots (x_n, y_n)\} \subset \mathcal{X} \times \mathcal{Y}$ drawn from an unknown data generating distribution $\mathcal{P} \sim \mathcal{X} \times \mathcal{Y}$.  
The goal of the \emph{black-box classifier} is to learn a hypothesis $h$ that 
minimizes the expected empirical risk over observed training distribution $\mathcal{D}$.

\begin{equation}
h^* = \arg \min_{h}  \mathbb{E}_{ (x , y) \in \mathcal{D}} \ell(h(x), y) \label{eq:ERM}
\end{equation}
where $\ell(\cdot)$ is classification loss function (e.g., cross-entropy between predicted and ground-truth labels), and $\hat{y} = h(x)$ is the corresponding predicted class label.

\spara{Black-box Classifier's Uncertainty:} The degree of uncertainty in a prediction can be measured by the Shannon entropy over the outcomes for any given test point. Higher entropy corresponds to higher uncertainty.
\begin{equation}
H[Y \vert X] = - \sum_{y \in \mathcal{Y}}P(y \vert x, \mathcal{D}) \log_2 P(y \vert x, \mathcal{D})
\end{equation}

The overall uncertainty corresponding to the predictive task denoted as $H[Y \vert X]$
encompasses uncertainty due to aleatoric, epistemic and model uncertainty \cite{hora1996aleatory, der2009aleatory, senge2014reliable}. 

\subsection{\bf \em Mapping Failure Scenarios to Uncertainties} Next, we give a brief introduction to the types of uncertainties -- aleatoric, epistemic and model uncertainty -- in a predictive task, and draw a connection between predictive uncertainties and common sources of failures in ML system.

\begin{figure}[tbh!]
	\centering	
	\begin{subfigure}{0.35\linewidth}
		\centering	
		\includegraphics[scale=0.39]{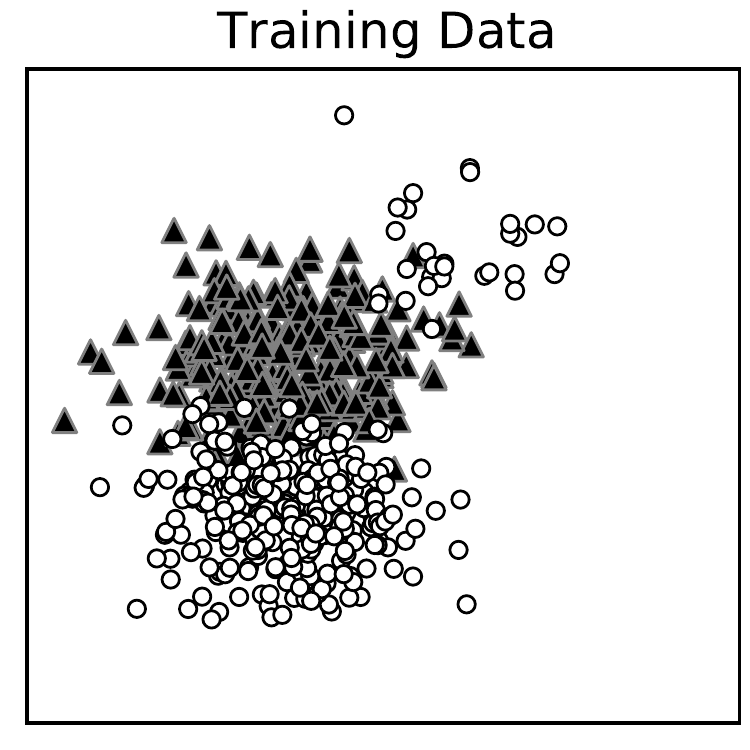}
		\vspace{-2mm}
		\caption{}
		\label{fig:synthetic-example-bbox-train-errors}
	\end{subfigure}
	\begin{subfigure}{0.63\linewidth}
		\centering	
		\includegraphics[scale=1.45]{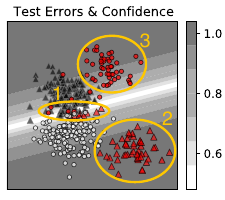}
		\vspace{-2mm}
		\caption{}
		\label{fig:synthetic-example-bbox-test-errors}
	\end{subfigure}	
	\caption{{\em Example}: (a) Training data for classification task (b) 
		learned decision boundary of an SVM classifier
		and different types of test-time errors, e.g., due to (1) data variability and noise (2) data shift, and (3) model limitations.}
	\label{fig:synthetic-example-bbox}
	\vspace{-4mm}
\end{figure}

\spara{Example:}
These different kinds of uncertainty are illustrated via a synthetic example in Fig. \ref{fig:synthetic-example-bbox}. 
We will use this as running example to motivate the proposed approach.
Consider the classification task dataset in Fig. \ref{fig:synthetic-example-bbox}. 
The position on x-axis and y-axis represents input features. 
The markers (black triangles and white circles) represent binary class labels. 
A linear SVM classifier, for example, would learn a decision boundary that best discriminates the two classes as shown in Fig. \ref{fig:synthetic-example-bbox-test-errors}. The test-errors made by the model are highlighted in red.

Firstly, in many predictive tasks $Y$ can rarely be estimated deterministically from $X$ due to inherent stochasticity in the dataset, 
a.k.a \emph{aleatoric} uncertainty.
For instance, errors arising due to inherent data variability and noise, marked as Region 1 in Fig. \ref{fig:synthetic-example-bbox-test-errors}).
Such errors are inherently \emph{irreducible} (unless additional features are collected).
Additionally, there is uncertainty arising due to ``lack of knowledge'' about the true data generating process.
For instance, consider the test errors caused by shifts in the data distribution, marked as Region 2.
Such errors due to {\em epistemic} uncertainty can in principle be mitigated by collecting additional training data and retraining the model.
Further, ML models have additional uncertainty in estimating the {true} model parameters given limited training data.
For instance, consider \emph{systematic} errors arising due to fitting a linear model to non-linear data, marked as Region 3. 
Errors due to \emph{model} uncertainty can in principle be addressed (e.g., by training a model from a different model class).%

\begin{figure}[tbh!]
	\centering	
	\begin{subfigure}{0.35\linewidth}
		\centering	
		\includegraphics[scale=0.134]{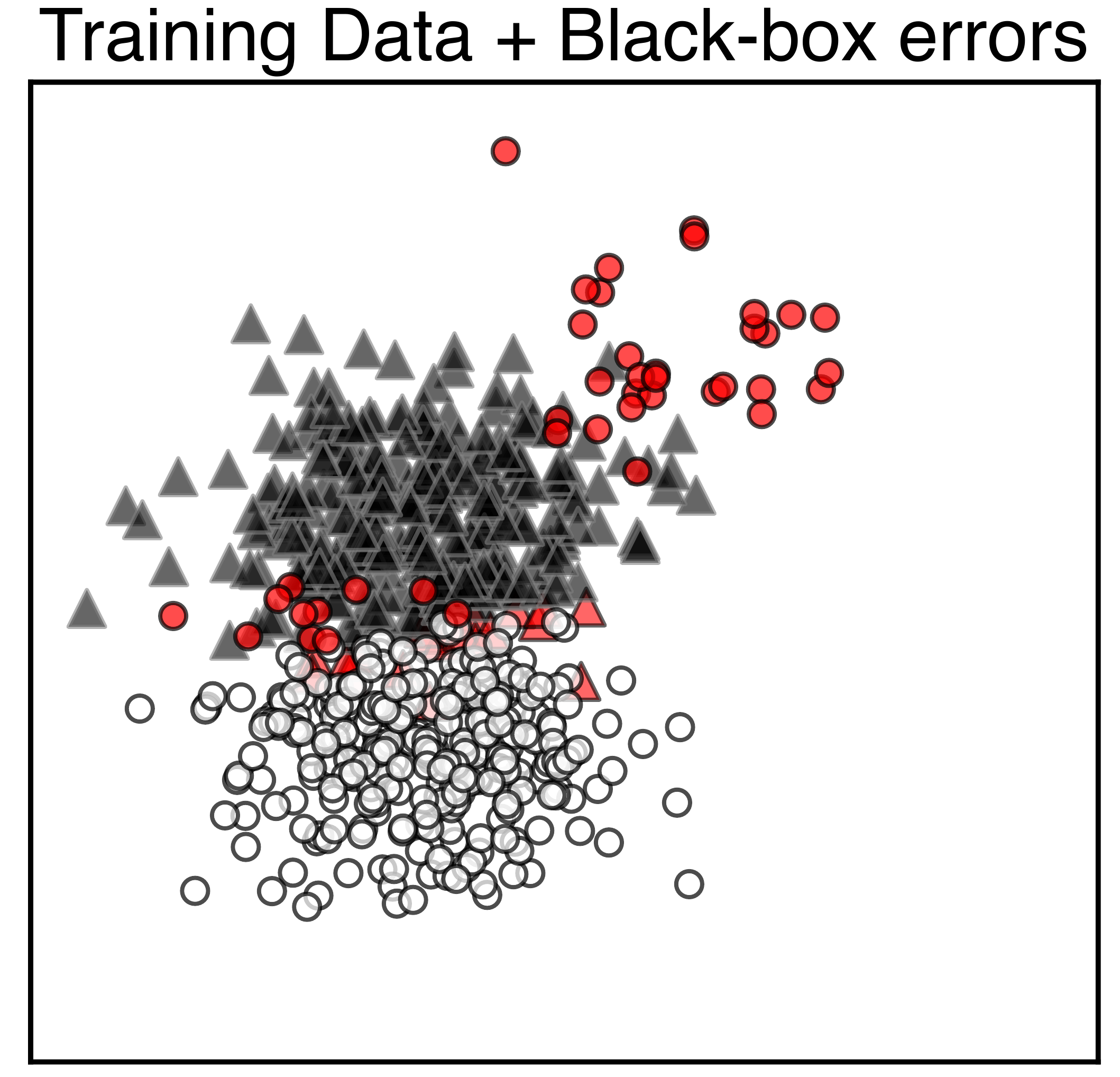}
		\caption{}
		\label{fig:synthetic-example-metalearner-input}
	\end{subfigure}
	\begin{subfigure}{0.63\linewidth}
		\centering	
		\includegraphics[scale=0.37]{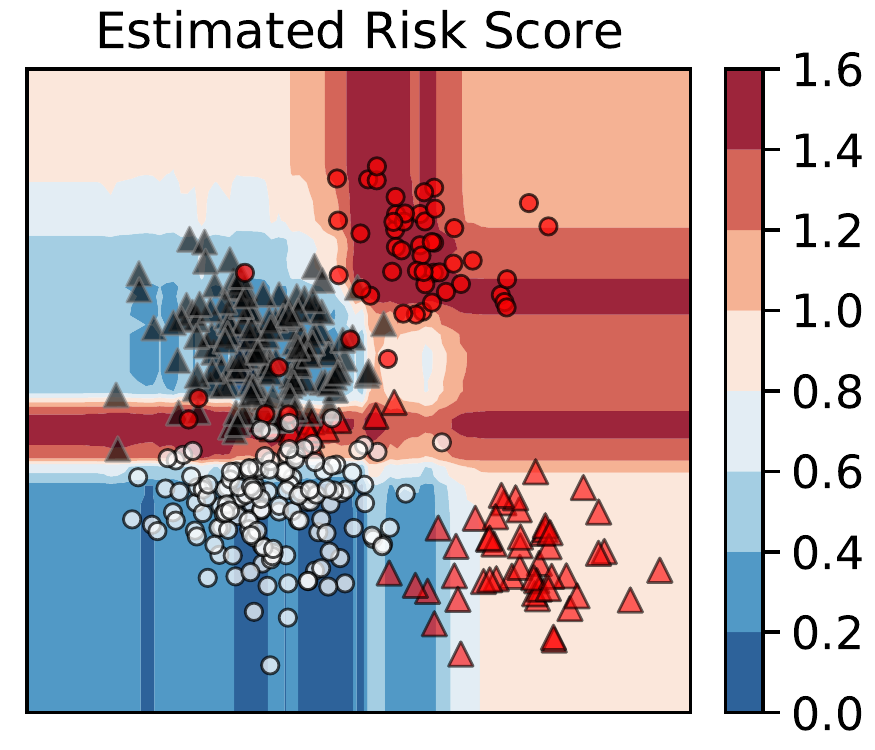}
		\caption{$P(z \vert x) +  \mathbb{E}_{f} [H[P(z \vert x, f)] + \mathcal{I}[z, f \vert x, \mathcal{D}]$}
		\label{fig:synthetic-example-metalearner-riskscore}
	\end{subfigure}
	\vspace{2mm}
	\newline
	\begin{subfigure}{0.3\linewidth}
		\centering	
		\includegraphics[scale=0.37]{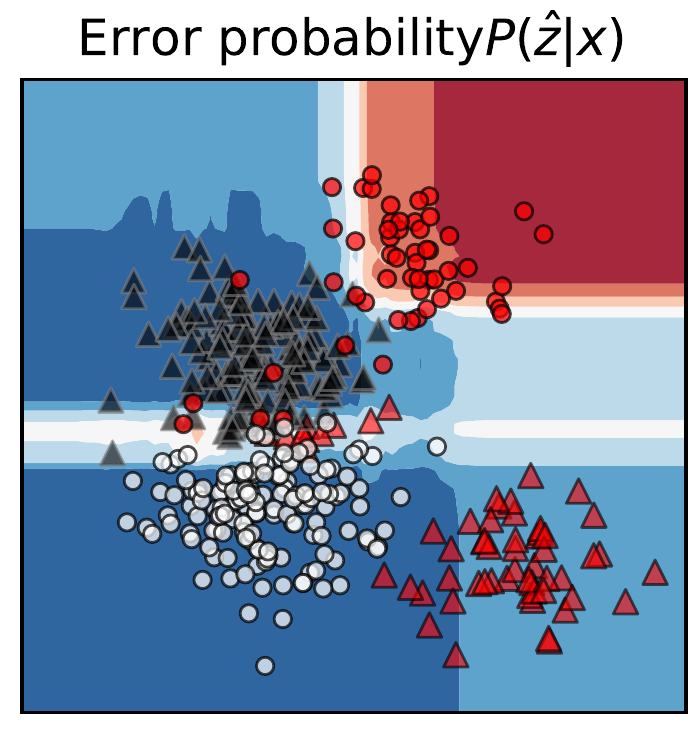}
		\caption{$P(z \vert x)$}
		\label{fig:synthetic-example-metalearner-error-prob}
	\end{subfigure}	
	\begin{subfigure}{0.3\linewidth}
		\centering	
		\includegraphics[scale=0.37]{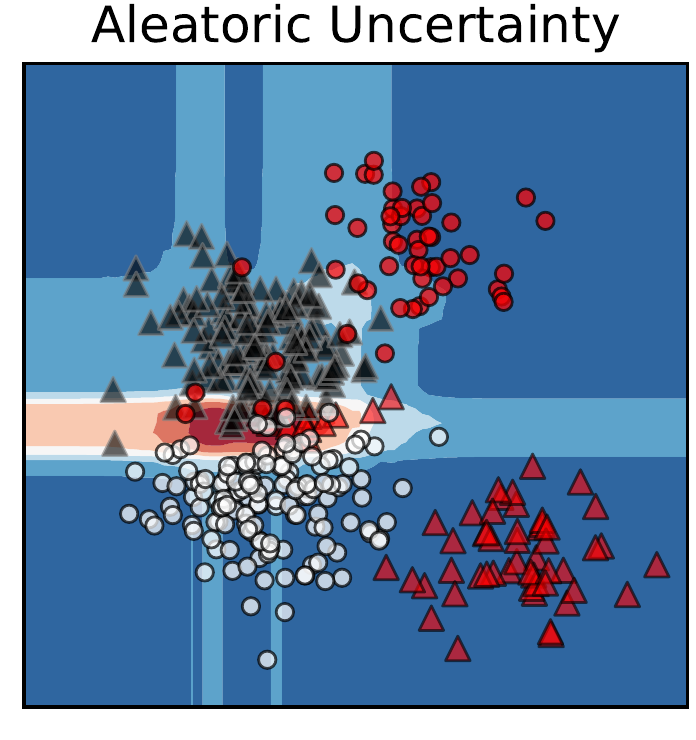}
		\caption{$\mathbb{E}_{f} [H[P(z \vert x, f)]$}
		\label{fig:synthetic-example-metalearner-alea-uncertainty}
	\end{subfigure}	
	\begin{subfigure}{0.37\linewidth}
		\centering	
		\includegraphics[scale=0.37]{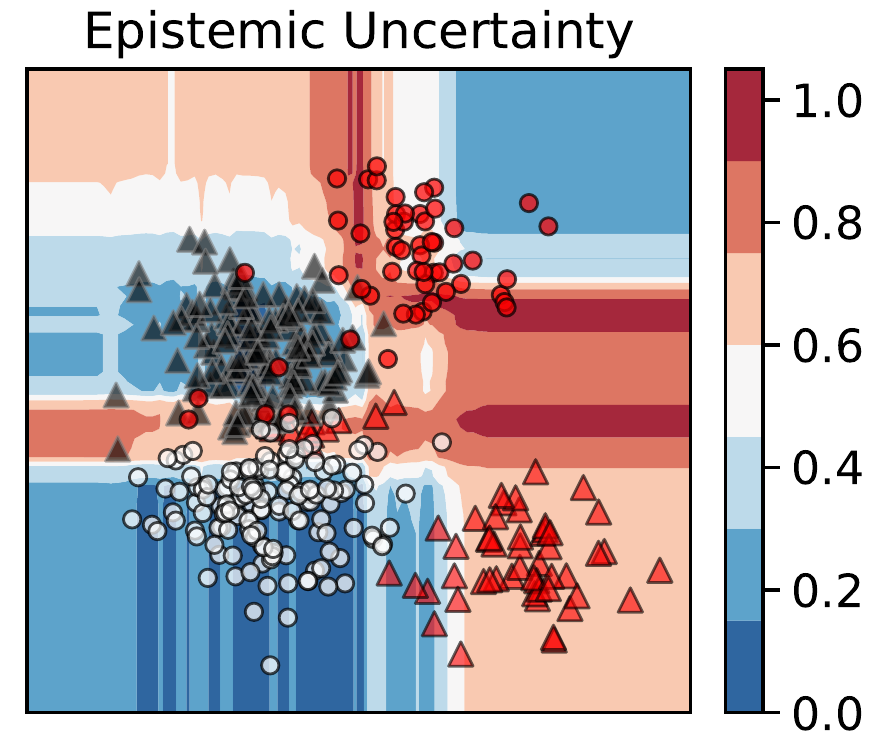}
		\caption{$\mathcal{I}[z, f \vert x, \mathcal{D}]$}
		\label{fig:synthetic-example-metalearner-epis-uncertainty}		
	\end{subfigure}
	\caption{Meta-learner: (a) Training input to \emph{meta-learner} (b) 
	\emph{meta-learner's} estimated overall {\em risk score} (c, d, e) decomposition of the overall risk score into its various constituting components, i.e., (c) model, (d) aleatoric and (e) epistemic uncertainty, that capture errors due to (c) model limitations, (d) data variability and noise, and (e) data shift, respectively.}
	\label{fig:synthetic-example-metalearner}
	\vspace{-3mm}
\end{figure}

\subsection{\bf \em Design Rationale}
We draw inspiration from \emph{Fano's Inequality}~\cite{fano1961transmission,cover1999elements}, a classic information-theoretic inequality which when viewed from a ML perspective
draws a connection between predictive uncertainty $H[Y \vert X]$, uncertainty in error prediction $H[Z \vert X]$, and probability of error $P(Z \vert X)$ of a Bayes optimal classifier,
where $Z$ is a random variable indicating prediction error
$Z := \mathbb{I}(Y \neq \hat{Y})$.

\smallskip
\noindent
\textbf{Fano's Inequality \cite{fano1961transmission, cover1999elements}}:
\emph{Consider
 random variables $X$ and $Y$, where $Y$ is related to $X$ by the joint distribution $P(x, y)$. Let $\hat{Y}=h(X)$ be an estimate of $Y$, with the random variable $Z$ representing an occurrence of error, i.e., $Z := \mathbb{I}(Y \neq \hat{Y})$. Fano's inequality states that
}
\begin{equation}
H[Y \vert X] \leq   H[Z \vert X] + P(Z \vert X) \cdot \log_2 (\norm{\mathcal{Y}}-1) \label{lemma:fanos-inequality} 
\end{equation}
where  $\norm{\mathcal{Y}}$ is the number of classes, $H$ is Shannon entropy,  and $P(Z \vert X)$ is probability of error.

\spara{Key Idea:} The conditional entropy $H[Z \vert X]$ and the error probability $P(Z \vert X)$ in Eq. \ref{lemma:fanos-inequality} are
not known, but we can 
approximate them by computing empirical estimates of conditional entropy $H_f[Z \vert X]$ and error probability $P_f(Z \vert X)$ of a separate \emph{meta-learner} $f: X \rightarrow Z$ whose goal is to predict errors $Z$ made by the underlying black-box classifier $h$ with respect to the original classification task.

Given such a meta-learner $f$, we argue that a 
black-box model's classification errors on unseen data, which relate to the uncertainty 
$H[Y \vert X]$, can be estimated by combining 
$f$'s
predicted probability of error $P_f(Z \vert X)$ and 
$f$'s
own uncertainty corresponding to predicting errors $H_f[Z \vert X]$.

\spara{Example:} 
Let us revisit the synthetic example of Fig. \ref{fig:synthetic-example-bbox},
looking at it from a meta-learner's perspective.
Fig. \ref{fig:synthetic-example-metalearner} shows
different perspectives on this setting.

Fig. \ref{fig:synthetic-example-metalearner-input} visualizes the input to the meta-learner, which consists of training datapoints $X$ and the black-box model's training errors $Z$ (highlighted in red).
Observe that errors due to \emph{model limitations} (top right red points) appear as systematic errors in the input space, and are \emph{predictable}.
We argue that by training a \emph{meta-learner} to predict 
black-box classification model's errors, we can capture these {systematic errors} due to model limitations with meta-learner's predicted {\em error probabilities} $P_f(Z \vert X)$, as shown in Fig. \ref{fig:synthetic-example-metalearner-error-prob}.

Further, recall that both aleatoric and epistemic uncertainties are related to the underlying training data.
We posit that the meta-learner, which is trained on the same data samples as the black-box classifier, inherits these data-induced uncertainties, and this is reflected in the meta-learner's 
{\em aleatoric} and {\em epistemic} uncertainties, as shown in
\ref{fig:synthetic-example-metalearner-alea-uncertainty} and Fig. \ref{fig:synthetic-example-metalearner-epis-uncertainty}.

The intuition is as follows.
Consider the region near the decision boundary in Fig. \ref{fig:synthetic-example-metalearner-input}.
As the meta-learner sees both \emph{failure} and \emph{success} cases of the black-box classifier in this region, the meta-learner, too, has \emph{aleatoric uncertainty} in this region of inherent noise.
Similarly, consider 
the test points situated far away from the training data.
The meta-learner would also have significant epistemic uncertainty in its error prediction,
 as it has not seen any training data in this region. 
Thus, by estimating the meta-learner's 
{\em own} {aleatoric and epistemic uncertainty}, we 
can indirectly 
capture the black-box classifier's {aleatoric} and {epistemic} uncertainty, as shown in Fig. \ref{fig:synthetic-example-metalearner-alea-uncertainty} and Fig. \ref{fig:synthetic-example-metalearner-epis-uncertainty},
respectively.
In our experiments, we will present empirical evidence of these insights.

\smallskip
\noindent
Putting these three insights together, 
we propose the combined notion of {\em risk score},
as shown in Fig. \ref{fig:synthetic-example-metalearner-riskscore}.
The estimated risk score (background color) is high in the regions of actual test-time errors.

In the following,
Subsection
\ref{subsec:model-bbox-uncertainty} 
formalizes the meta-learner's task, and presents our proposed
meta-learner ensemble
for the {Risk Advisor}.
Subsection
\ref{subsec:model-epis-alea} discusses how to refine the overall uncertainty into informative components for different kinds of uncertainty, and 
compute overall
\emph{risk score}.

\subsection{\bf \em Meta-learner Ensemble}\label{subsec:model-bbox-uncertainty}

\vspace{-1mm}
\spara{Meta-learner's Task:}
Given input training samples $x \in {X}$, predicted class labels $\hat{y}:=h(x)$ of a fully trained \emph{black-box classifier} $h$, and a random variable $Z := \mathbb{I}(Y \neq \hat{Y})$ indicating errors of the \emph{black-box classifier} $h$ with respect to the original classification task. 
Our goal is to learn an meta-learner $f: X \rightarrow Z$ trained to predict errors of the \emph{black-box classifier} with respect to the original task given by
\begin{equation}
f = \arg \min_{f \in \mathcal{F}}  \mathbb{E}_{ (x , z) \in \mathcal{D}} \ell(f(x), z) \label{eq:meta-learner}
\end{equation}
where 
$z$ is a random variable indicating errors of the base classifier predictor given by $z =  \mathbb{I}(y \neq \hat{y})$,
$\ell$ is a classification loss function. 
Given a newly seen test point $x^*$, the \emph{meta-learner's} predicted probability of error is given by $P(z \vert f, x^*)$.

However, the probability of error $P(z \vert f, x^*)$ estimated by
 a single
meta-learner $f$ can be biased due to its own uncertainty in the model parameters $P(f \vert \mathcal{D})$.
Next, we show how we can obtain a reliable estimate of the black-box model's
error probability
by training an ensemble of $M$ independent stochastic gradient boosted trees $\mathcal{F} = \{ P(z \vert x^*, f^m )\}_{m=1}^{M}$, 
and computing their expectation.

\spara{Ensemble of Stochastic Gradient Boosted Trees (E-SGBT):} 

We consider an ensemble of $M$ independent models $\mathcal{F} = \{ f^m \}_{m=1}^{M}$ such that each of the individual models $f^m$ is 
a stochastic gradient boosted tree (SGBT)~\cite{friedman2002stochastic}. 
Note that the proposed \emph{E-SGBT} is an ensemble of ensembles, i.e., each of the $M$ SGBT's in the ensemble is itself
an ensemble of $T$ weak learners
trained iteratively via bootstrap aggregation.
To ensure minimum correlation between the $M$ individual models in our ensemble, we introduce randomization in two ways.
First, each of the SGBTs in the ensemble is initialized with a different random seed. 
Second,
each of the individual 
SGBTs
is itself an ensemble of $T$ weak learners trained iteratively via bootstrap aggregation.
Specifically, for each SGBT in the \emph{E-SGBT} ensemble, at each iteration, a subsample of training data of size $\tilde{N}<N$  is drawn at random, without replacement, from the full training dataset. The fraction $\frac{\tilde{N}}{N}$ is called the sample rate. The smaller the sample rate, the higher the difference between successive iterations of the weak learners, thereby introducing randomness into the learning process. 

Given $M$ error probability estimates $\{P(z \vert x, f^m)\}_{m=1}^{M}$ 
by each of the models in the ensemble, an estimate of the probability of error $P(z \vert x , \mathcal{D})$ can be computed by taking the expectation over all the models in the ensemble: 
\begin{equation}
P(z \vert x , \mathcal{D}):= \mathbb{E}_{f \in \mathcal{F}}[P(z \vert x, f, \mathcal{D})] \approx \frac{1}{M} \sum_{m=1}^{M}P(z \vert x, f^m,\mathcal{D}) \label{eq:error-prob}
\end{equation} 

The total uncertainty in the error prediction $H[P(z \vert x , \mathcal{D})]$ 
can be computed as the Shannon entropy corresponding to the estimated probability of error 
\begin{equation}
H[P(z \vert x , \mathcal{D})] = - \sum_{z \in \mathcal{Z}}P(z \vert x, \mathcal{D}) \log_2 P(z \vert x, \mathcal{D}) \label{eq:total-uncertainty}
\end{equation}

\subsection{\bf \em Identifying Sources of Uncertainty}  \label{subsec:model-epis-alea}
To distinguish between different sources of uncertainty -- data variariability/noise vs. data shifts between training and deployment data --
we compute
estimates of the \emph{aleatoric} and \emph{epistemic} uncertainty given an ensemble of $M$ independent stochastic gradient boosted trees  $\mathcal{F} = \{ f^m \}_{m=1}^{M}$.
This approach was originally
developed in the context of neural networks~\cite{depeweg2018decomposition}, 
but the idea is more general and has recently been applied using ensembles of gradient boosted trees and random forests~\cite{ustimenko2020uncertainty, shaker2020aleatoric}.

\spara{Decomposing Aleatoric and Epistemic Uncertainty:} 
The main idea is that in the case of data points with epistemic uncertainty (e.g., out-of-distribution points), the $M$ independent models in the ensemble given $\mathcal{F}:=\{ f^m \}_{m=1}^{M}$ are likely to yield 
a diverse set of predictions (i.e., different output labels) for similar inputs.
In contrast, for data points with low epistemic uncertainty (e.g., in-distribution points in dense regions), they are likely to agree in their predictions. 
Hence, by fixing $f$,
 the 
\emph{epistemic} uncertainty can be removed, and  the \emph{aleatoric} uncertainty can be computed by 
taking the expectation
over all models $f \in \mathcal{F}$.
\begin{equation}\label{eq:alea-uncertainty-analytical}
\mathbb{E}_{p(f \vert \mathcal{D})} H[P(z \vert x, f)] = \int_{\mathcal{F}} P(f \vert D) \cdot H[P(z \vert f, x)] df
\end{equation}

\spara{Aleatoric Uncertainty:} Given $M$ predicted probability estimates $\{P(z \vert x, f^m)\}_{m=1}^{M}$ for each of the models in the ensemble, an estimate of \emph{aleatoric uncertainty} in Eq. \ref{eq:alea-uncertainty-analytical} 
can be empirically approximated by averaging over individual models $f^m \in \mathcal{F}$ in our \emph{E-SGBT} ensemble. 
\begin{equation}\label{eq:alea-uncertainty-empirical}
\mathbb{E}_{f \in \mathcal{F}} [H[P(z \vert x, f)]] \approx \frac{1}{M} \sum_{m=1}^{M}  H[P(z \vert x, f^m)]
\end{equation}

\spara{Epistemic uncertainty:} Finally, \emph{epistemic uncertainty} can be computed as the difference between \emph{total uncertainty} and \emph{aleatoric uncertainty}.
\begin{equation}\label{eq:epis-uncertainty-empirical}
\small
\underbrace{\mathcal{I}[z, f \vert x, \mathcal{D}]}_{\text{Epistemic Uncertainty}} = \underbrace{H[P(z \vert x , \mathcal{D})]}_{\text{Total Uncertainty}} - \underbrace{\mathbb{E}_{f \in \mathcal{F}} [H[P(z \vert x, f)]]}_{\text{Aleatoric Uncertainty}}
\end{equation}
where \emph{total uncertainty} is the entropy corresponding to the estimated probability of error $P(z \vert x , \mathcal{D})$
given in Eq. \ref{eq:total-uncertainty}.

\spara{Risk Score:} 
Putting it all together, our proposed \emph{Risk Score}, which captures black-box model errors arising due to all sources of uncertainty, can be computed as the sum of (i) predicted probability of error assigned by the meta-learner, i.e., model uncertainty, (ii) epistemic uncertainty and (iii) aleatoric uncertainty.
\begin{align}\label{eq:est-uncertainty-score}
\small
\text{Risk Score} & :=  \underbrace{P(z \vert x , \mathcal{D})}_{\text{Error probability}} +  \underbrace{H[P(z \vert x)]}_{\text{Total uncertainty}} \\ \nonumber
& = \underbrace{P(z \vert x , \mathcal{D})}_{\text{Model Uncertainty}} + \underbrace{\mathcal{I}[z, f \vert x, \mathcal{D}]}_{\text{Epistemic uncertainty}} +  \underbrace{\mathbb{E}_{f} [H[P(z \vert x, f)]}_{\text{Aleatoric uncertainty}}
\end{align}
Note that this \emph{risk score} 
is neither a probability nor an entropy measure, but it proves to be
a very useful indicator for failure risks in our experiments.
One could consider a weighted sum of each of the components to account for associated \emph{risk costs} for each type of error. For instance, if a system designer had expert knowledge that errors due to distribution shift (i.e., epistemic uncertainty) are more harmful,
she could assign more weight to the \emph{epistemic uncertainty} component.
In our experiments we assign equal weights.

\spara{Inference:} 
The meta-learner is trained on the underlying base-classifier's training data.
Given a newly seen test point $x^*$ at deployment-time, 
the \emph{Risk Advisor} computes predicted error probabilities for each of the $M$ models in the E-SGBT ensemble $\{P(z \vert x^*, f^m)\}_{m=1}^{M}$.
These values are fed into the 
\emph{Risk Advisor's}
estimated
\emph{error probability} in Eq. \ref{eq:error-prob},
aleatoric uncertainty in Eq. \ref{eq:alea-uncertainty-empirical}, epistemic uncertainty in Eq. \ref{eq:epis-uncertainty-empirical}
and risk score in Eq. \ref{eq:est-uncertainty-score}. 
Note that at deployment-time, we only expect the unseen data point $x^*$, and the trained 
meta-learner.

\section{Synthetic-Data Experiments}\label{sec:synthetic-experiments}
In this section,
we evaluate 
\emph{Risk Advisor's}
ability to
detect the sources of uncertainty inducing the failure risk.
To this end, we systematically generate
synthetic datasets covering a variety of ML failure scenarios including errors due (i) 
black-box classifier's model limitations (e.g., applying a linear model to non-linear decision boundary) (ii) data shift and (iii)  inherent data variability and noise. 
We then evaluate if the {\em Risk Advisor's} estimates for \emph{model}, \emph{epistemic}, and \emph{aleatoric} uncertainty can correctly capture the corresponding test-time errors made by the black-box classification model.

\begin{figure}[tbh]
	\begin{subfigure}{0.325\linewidth}
		\centering	
		\includegraphics[scale=0.31]{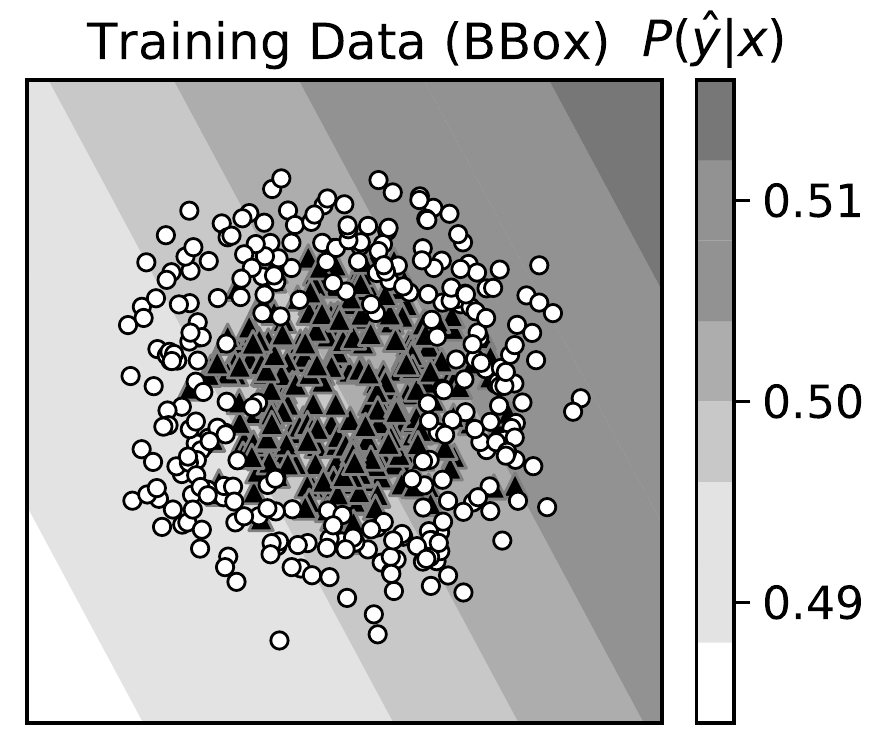}
		\vspace{-2mm}%
		\caption{}
		\label{fig:model-errors-training_data}
	\end{subfigure}
	\begin{subfigure}{0.325\linewidth}
		\centering	
		\includegraphics[scale=0.31]{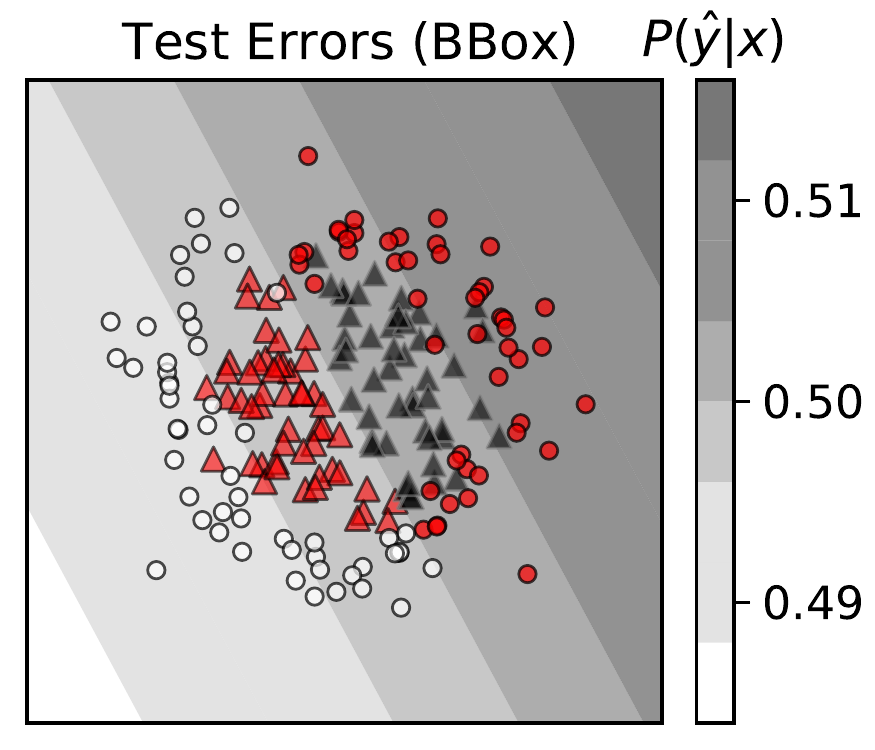}
		\vspace{-2mm}%
		\caption{}
		\label{fig:model-errors-test_data}
	\end{subfigure}
	\begin{subfigure}{0.325\linewidth}
		\centering	
		\includegraphics[scale=0.31]{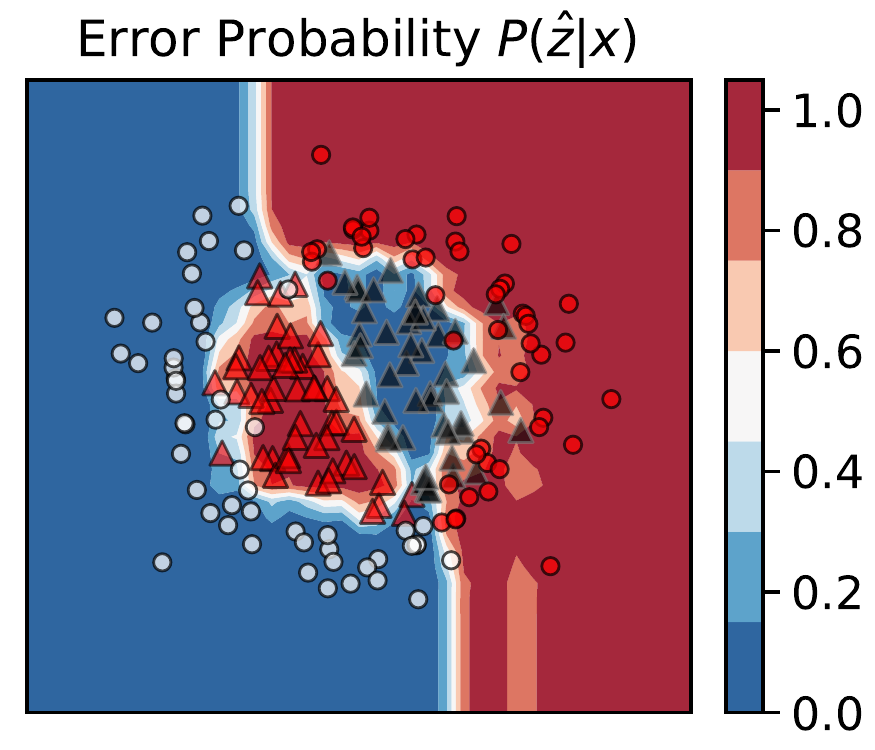}
		\vspace{-2mm}%
		\caption{}
		\label{fig:model-errors-pred_error}
	\end{subfigure}			
	\vspace{-2mm}
	\caption{Errors from model limitations: \emph{Risk Advisor's} estimated error probability $P(z \vert x)$ correctly identifies errors due to model limitation.}
	\label{fig:model-errors-synthetic-dataset}
	\vspace{-3mm}
\end{figure}

\spara{Errors due to Black-box Classifier's Model Limitations:} 
In order to simulate this scenario,
we construct a classification dataset with a non-linear decision boundary, i.e., two concentric circles \cite{scikit-learn}. We then fit a misspecified classification model to the task, i.e., a logistic regression classifier with a (log-)linear decision boundary as shown in Fig. \ref{fig:model-errors-synthetic-dataset}. 
The contour plot in Fig. \ref{fig:model-errors-training_data} visualizes the training data
and the learned decision boundary.
Fig. \ref{fig:model-errors-test_data} visualizes the test data. Test-set errors made by the black-box model are highlighted in red.

The contour plot in Fig. \ref{fig:model-errors-pred_error} visualizes the \emph{Risk Advisor's} predicted \emph{Error probability} ($P(\hat{z} \vert x)$).
Ideally, we would expect the \emph{Risk Advisor} to assign a higher error score for regions of the input space where the black-box classifier makes errors due to its model limitations.
We clearly see this trend: the \emph{Risk Advisor} correctly identifies the regions where the 
black-box classifier is likely to make errors due to its incorrect linear decision boundary.
This is especially remarkable given that the \emph{Risk Advisor} has no knowledge of the model family of the underlying black-box model (e.g., whether it is log-linear model or a neural network). In spite of having no information about the underlying model (other than its predictions), the \emph{Risk Advisor} is able to correctly capture the \emph{model uncertainty}. 

\begin{figure}[tbh]
	\begin{subfigure}{0.325\linewidth}
		\centering	
		\includegraphics[scale=0.31]{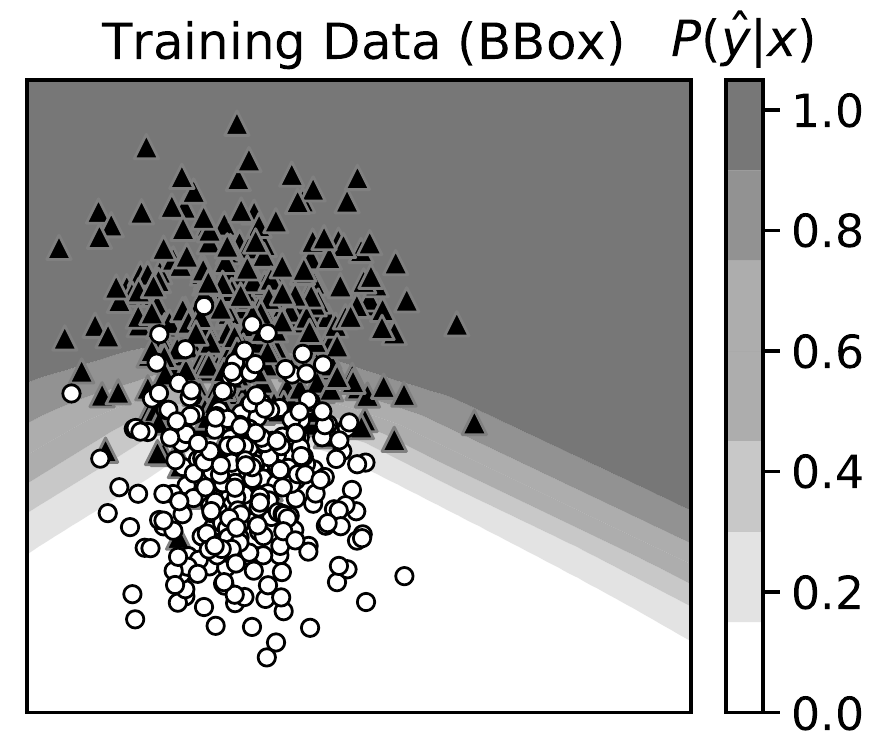}
		\vspace{-2mm}%
		\caption{}
		\label{fig:GMM-training_data}
	\end{subfigure}
	\begin{subfigure}{0.325\linewidth}
		\centering	
		\includegraphics[scale=0.31]{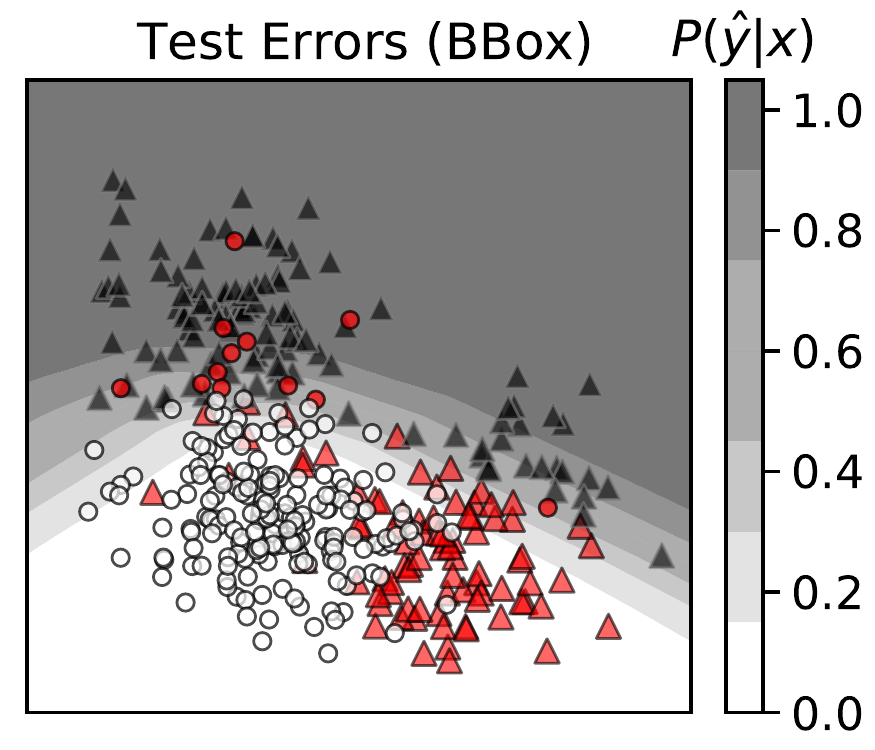}
		\vspace{-2mm}%
		\caption{}
		\label{fig:GMM-test_data}
	\end{subfigure}
	\begin{subfigure}{0.325\linewidth}
		\centering	
		\includegraphics[scale=0.31]{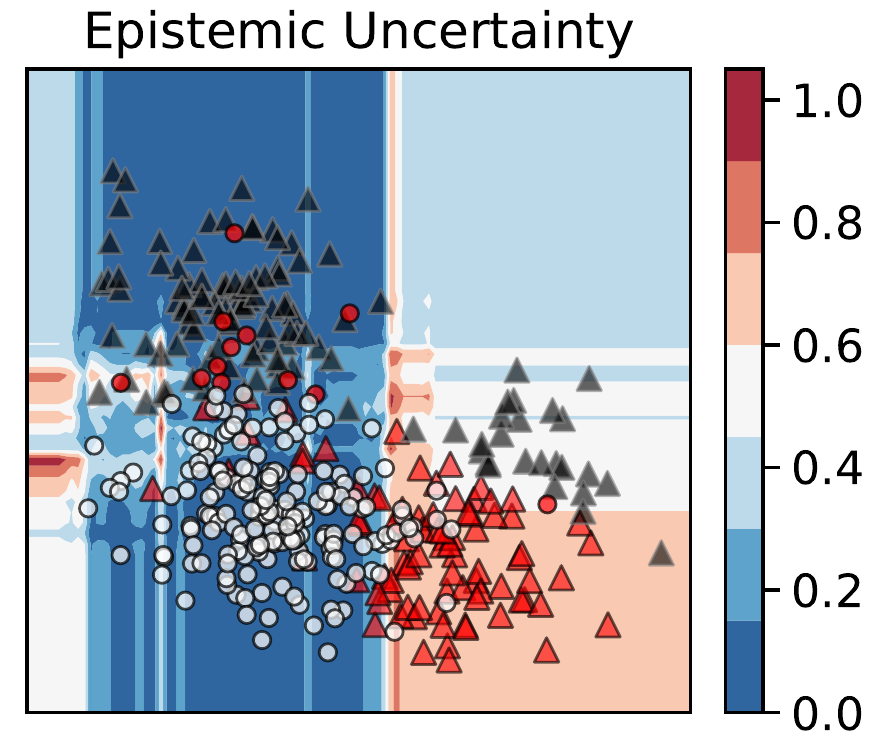}
		\vspace{-2mm}%
		\caption{}
		\label{fig:GMM-epis}
	\end{subfigure}
	\vspace{-2mm}%
	\caption{Errors from distribution shift:  Risk-Advisor's epistemic uncertainty correctly identifies test points far away from training distribution.}
	\label{fig:GMM-synthetic-dataset}
	\vspace{-2mm}
\end{figure}

\spara{Errors due to Distribution Shift:} 
In order to simulate a distribution shift scenario, we draw points from a mixture of two Gaussians. 
For the training points we set the mixture coefficient for one of the Gaussians to zero;
for the test points both mixture components are active.
This way, we are able to construct a dataset containing out-of-distribution test points as shown in Fig. \ref{fig:GMM-synthetic-dataset}. %
Fig. \ref{fig:GMM-training_data} visualizes the training data and the decision boundary learned by
a 2-layer feed-forward neural network (NN). 
Fig. \ref{fig:GMM-test_data} visualizes the test data.
Test errors of the NN are highlighted in red.
Observe that the NN \emph{misclassifies} out-of-distribution test points while (incorrectly) reporting high confidence.
The contour plot in Fig. \ref{fig:GMM-epis} visualizes \emph{Risk Advisor's} estimated \emph{epistemic} uncertainty.  

Ideally, we would like to see that the \emph{epistemic} uncertainty increases as we move towards the sparse regions of the training data, and that it is high for out-of-distribution regions. 
Despite some noise, we clearly see this trend: regions of low epistemic uncertainty (i.e., dark-blue regions) coincide with the dense in-distribution test points. \emph{Epistemic} uncertainty increases as we move towards sparse regions, and the values are especially high for out-of-distribution regions (bottom right in Fig. \ref{fig:GMM-epis}).

\vspace{-1mm}

\begin{figure}[tbh!]
	\begin{subfigure}{0.325\linewidth}
		\centering	
		\includegraphics[scale=0.31]{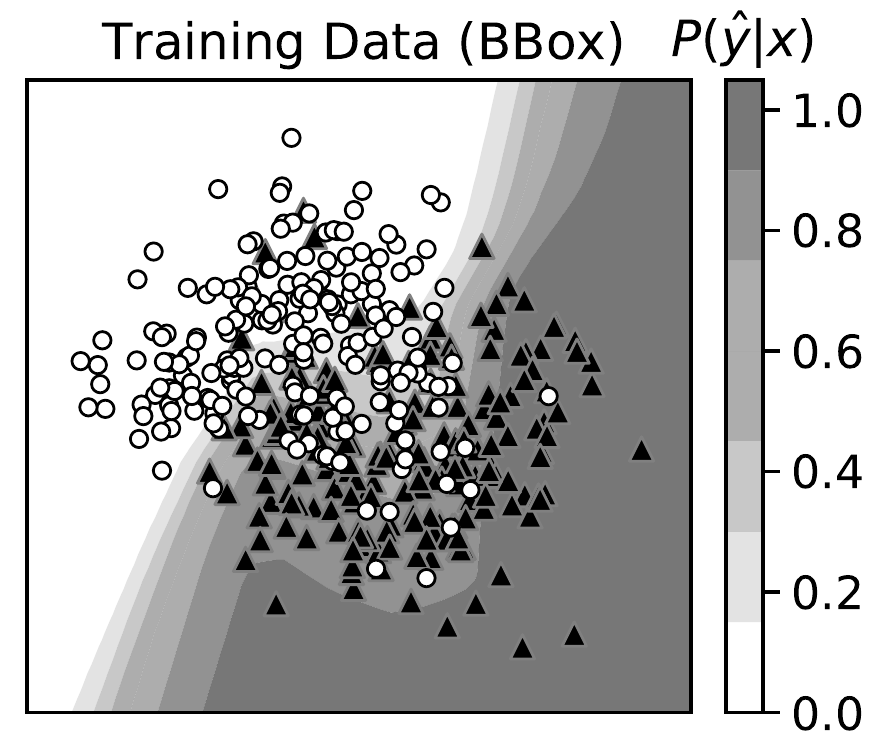}
		\vspace{-2mm}%
		\caption{}
		\label{fig:moon-training_data}
	\end{subfigure}
	\begin{subfigure}{0.325\linewidth}
		\centering	
		\includegraphics[scale=0.31]{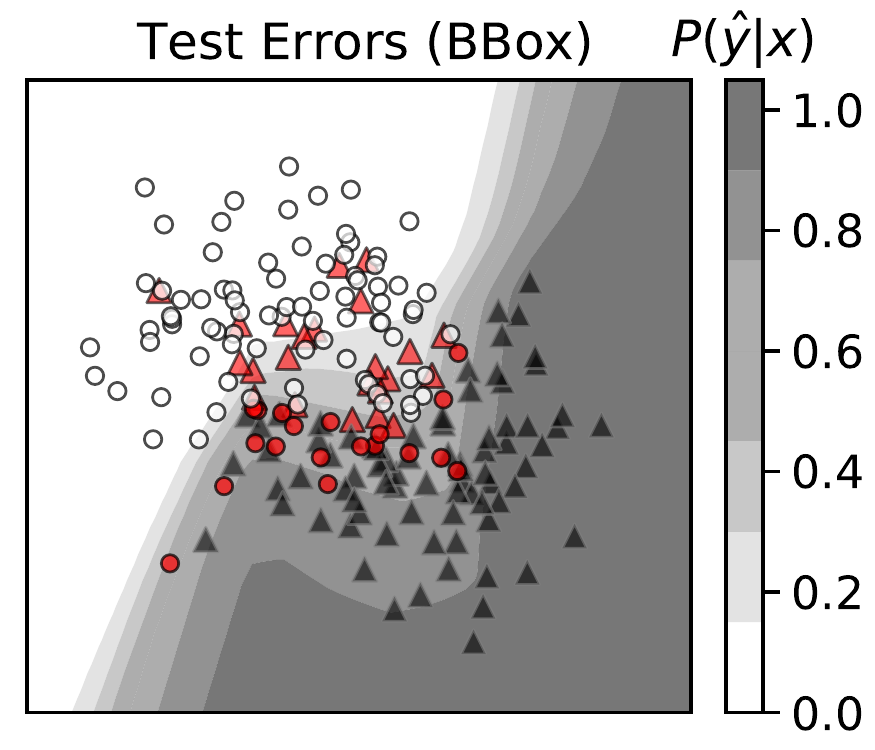}
		\vspace{-2mm}%
		\caption{}
		\label{fig:moon-test_data}
	\end{subfigure}
	\begin{subfigure}{0.325\linewidth}
		\centering	
		\includegraphics[scale=0.31]{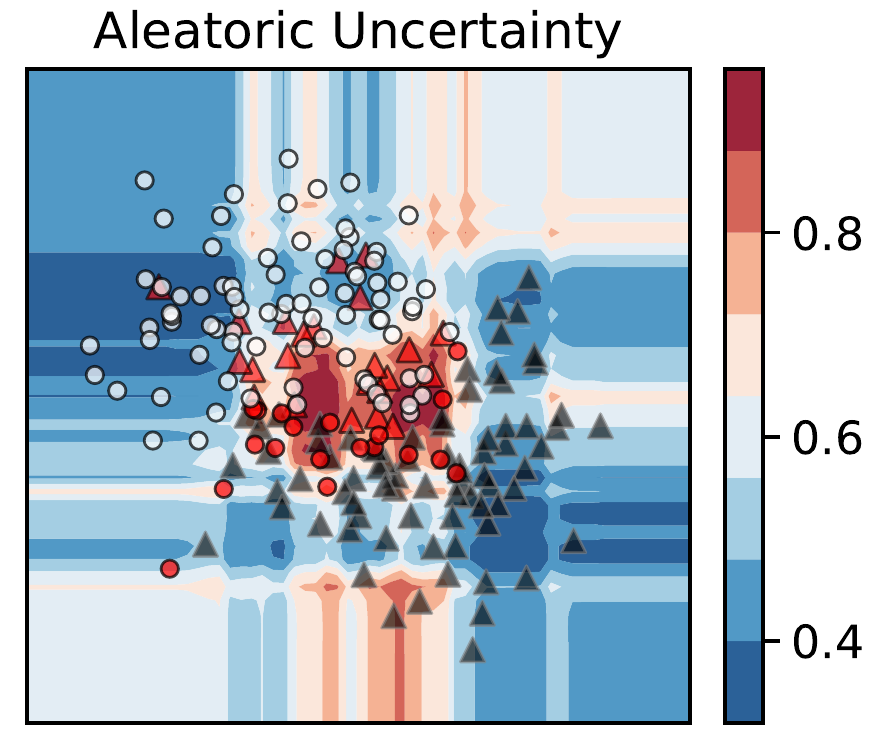}
		\vspace{-2mm}%
		\caption{}
		\label{fig:moon-alea}
	\end{subfigure}	
	\vspace{-2mm}
	\caption{Errors from data variability and noise: Risk Advisor's aleatoric uncertainty correctly identifies test points in the regions 
		with class overlap.}
	\label{fig:Moon- synthetic-dataset}
	\vspace{-2mm}
\end{figure}

\spara{Errors due to Data Variability and Noise:} 
To simulate a dataset with inherent noise, we draw points from the classic 
two-moons dataset \cite{scikit-learn}, and add Gaussian noise with standard deviation $0.5$ to the dataset as shown in Fig. \ref{fig:Moon- synthetic-dataset}. 
Fig. \ref{fig:moon-training_data} visualizes the training data and the decision boundary learned by a 2-layer feed-forward neural network (NN). 
Fig. \ref{fig:moon-test_data} visualizes the test data. Test-errors are highlighted in red.

The contour plot in Fig. \ref{fig:moon-alea} visualizes estimated \emph{aleatoric} uncertainty.
Ideally, we would expect
that
\emph{aleatoric} uncertainty is high for the regions with large class overlap.
We clearly see this trend: the estimated \emph{aleatoric} uncertainty is high for the test points near the decision boundary,
with high class overlap.

\vspace{-1mm}

\section{Real-World-Data Experiments}\label{sec:real-experiments}
In this section, we evaluate the performance of {\em Risk Advisor} by performing extensive experiments on 9 real-world datasets
and on 6 families of black-box classification models.
First, we evaluate the \emph{Risk Advisor's} ability to {\em predict failure risks} at deployment time (Subsection \ref{subsec:detect-errors}).
Next, we investigate its performance on a variety of applications for {\em risk mitigation}, including (i) selectively abstaining under uncertainty (Subsection \ref{subsec:abstain-from-predictions})
(ii) detecting out-of-distribution test examples (Subsection \ref{subsec:detect-source-of-uncertainty}) 
and (iii) mitigating risk by judiciously collecting additional samples for re-training the system (Subsection \ref{subsec:sample-retrain}).

\subsection{\bf \em Experimental Setup} \label{subsec:experimental-setup}

\smallskip
\noindent
\textbf{Datasets:} %
We perform our evaluation on the following small and large benchmark classification datasets covering a variety of common ML failure scenarios:

\smallskip
\noindent
\emph{High-dimensional Image Datasets:}
\squishlist
\item \emph{CIFAR 10:} The CIFAR-10 dataset~\cite{cifar10} consists of 60K color images in 10 classes, including blurred and noisy images, which are specially prone to model failures.
\item \emph{MNIST:} The MNIST dataset~\cite{lecun2010mnist} consists of 60K grayscale images of handwritten digits in 10 classes. Due to the variability in writing style, certain images are prone to misclassification.
\item \emph{Fashion MNIST:} The fashion MNIST dataset~\cite{fashion-mnist} consists of 60K images of clothing and accessories in 10 classes, including images with rare and unusual product designs, which can be prone to errors.

\squishend

\smallskip
\noindent
\emph{Mission-critical Fairness Datasets:}
\squishlist
\item \emph{Census Income:} Recent work in ML fairness has shown that models often make more errors for underrepresented groups in training data. To 
simulate this setting, we consider the Adult dataset~\cite{uci}, a benchmark dataset in fairness literature, consisting of 49K user records.
The dataset contains underrepresented groups (e.g., Female). 
\item \emph{Law School:} Similarly, we use the LSAC dataset \cite{wightman1998lsac}
consisting of 28K law school admission records. The classification task is to predict whether a candidate would pass the bar exam. The dataset contains underrepresented groups (e.g., ``Black''). 
\squishend

\smallskip
\noindent
\emph{Distribution shift, unseen demographics/regions/domain:}
\squishlist
\item \emph{Census Income (Male $\rightarrow$ Male, Female):} To simulate 
distribution shift, we take the aforementioned \emph{Census Income} dataset, and exclude \emph{female} points from the training set. Our test set consists of both Male and Female points.
\item \emph{Law School (White $\rightarrow$ White, Black):} 
Similarly, we take the aforementioned \emph{Law School} dataset, and exclude user records from the \emph{Black} demographic group from the training set. The test set consists of both White and Black points.
\item \emph{Heart Disease:} A common ML failure scenario is when a ML model is applied to a new geographic region. To simulate this scenario we combine four different heart disease datasets available in the UCI repository~\cite{uci} by using a subset of features overlapping between them. 
We use the US Cleveland heart disease dataset
as our training dataset, and use it to predict heart disease on a UK statlog dataset, Hungarian (HU)
and Switzerland (CH) heart disease dataset.
\item\emph{Wine Quality:} Another failure scenario is when a
trained model 
is applied to
an application domain
for which it has inadequate or bad training data. To simulate this scenario, 
we train models on white wine, and apply it to predict quality of red wine in UCI wine dataset~\cite{uci}.
The classification task is to predict if the wine quality is $\geq 6$.
\squishend

\smallskip
\noindent
\textbf{Black-box Classification Models:} 
To demonstrate the versatility of \emph{Risk Advisor}, we evaluate it on classifiers from 6 different families, including deep neural models such as ResNet50 and CNN for the high dimensional image dataset, and 
classic ML algorithms such as SVM, Random Forests, Multi-layer Perceptron, and logistic regression
for tabular datasets.
Following are the implementation details:
\squishlist
	\item ResNet 50: The 50-layer deep residual network architecture~\cite{resnet50} trained with batch size of 128 for 100 epochs.
	\item CNN: A convolutional neural net with 2 convolutional layers with 32, 64 hidden units, max pooling, and ReLu activations, trained with batch size 128 for 10 epochs.
	\item MLP: Multi-layer perceptron with 2 hidden layers with $32, 16$  hidden units, batch size $64$, and ReLu activations.
	\item SVM: support vector machines with RBF kernel and Platt scaling~\cite{platt1999probabilistic} to produce probability estimates.%
	 \item RF: a random forest with $1000$ decision trees, bootstrap sampling, and max-features set to 'sqrt'. 
	 \item LR: logistic regression with L2 regularization. 
\squishend

\smallskip
\noindent
\textbf{State-of-the-art Baselines: } 
Our baseline comparison 
includes
the underlying black-box classification model's own (self-) {\em confidence scores}. 
Specifically, for all deep neural models, i.e., ResNET50, MLP, and CNN, we rely on the {confidence} score given by max class probability (DNN-MCP), as proposed by \cite{hendrycks2016baseline}, which is a well established strong baseline.
For RF's, we rely on the {\em uncertainty} score, computed as per the state-of-the-art method for random forest (RF-uncertainty), as proposed by \cite{shaker2020aleatoric}.
For SVM, we rely on the standard approach of computing {confidence} scores over prediction probabilities from decision values after Platt scaling (SVM-Platt)~\cite{platt1999probabilistic}.
For LR,  the {confidence} score is given by the distance from the decision boundary
(LR-Confidence).

Our main comparison is with the state-of-the-art method \emph{Trust Score} \cite{jiang2018trust}.
Similar to \emph{Risk Advisor}, Trust Score is 
a model-agnostic post-hoc approach, which takes as input a black-box classifier's predictions, and training data to produce point-wise {trust scores} for newly seen test points.

While calibrating a classifier's scores is a popular technique for producing calibrated confidence values, such techniques are rank-preserving. 
As all our evaluation metrics, i.e., AUROC, AUPR, and PRR (introduced later in
Subsections \ref{subsec:detect-errors}, \ref{subsec:detect-source-of-uncertainty} and \ref{subsec:abstain-from-predictions}) are based on seeing different relative rankings of the scores rather than absolute values, there is no point in comparing against 
rank-preserving calibration techniques.

\smallskip
\noindent
\textbf{Implementation:}
The {\em Risk Advisor}
is implemented as an ensemble of 10 SGBT classifiers,
each initialized
with a different random seed. 
Train and test sets are constructed using a 70:30 stratified split.
All categorical features are one-hot encoded.
Best hyper-parameters
are chosen via grid-search by performing 5-fold cross validation.
For {\em Risk Advisor's} E-SGBT model, we tune max-depth in [3,4,5,6], sample-rate in [0.25, 0.5, 0.75] and num-estimators in [100, 1000].
For \emph{Trust Score}, we use the code shared by \cite{jiang2018trust} and perform grid search over the parameter space reported in the paper.
All experiments are conducted
using scikit-learn and Keras on 2 GPUs and CPUs.
Results reported are mean values over 5 runs.

\begin{table*}[tbh!]
	\centering
	\setlength{\tabcolsep}{3pt}
	\resizebox{\textwidth}{!}{
		\begin{tabular}{l|c|c|c|cccc|cccc|cccc|cccc|cccc|cccc}
			\toprule
			Dataset &  CIFAR 10 & Fashion &   MNIST & \multicolumn{4}{c|}{Census Income} & \multicolumn{4}{c|}{Law School} & \multicolumn{4}{c|}{Wine Quality} & \multicolumn{4}{c|}{Heart Disease} & \multicolumn{4}{c|}{Census Income} & \multicolumn{4}{c}{Law School} \\
			&   & MNIST &    & \multicolumn{4}{c|}{} & \multicolumn{4}{c|}{} & \multicolumn{4}{c|}{white $\rightarrow$ white, red} & \multicolumn{4}{c|}{US $\rightarrow$ US, UK, CH, HU} & \multicolumn{4}{c|}{Male $\rightarrow$ Male, Female} & \multicolumn{4}{c}{White $\rightarrow$ White, Black} \\
			\midrule
			Black-box (BBox)& ResNet50 & CNN & CNN &        LR &   MLP &    RF &   SVM &         LR &   MLP &    RF &   SVM &    LR &   MLP &    RF &   SVM &    LR &   MLP &    RF &   SVM &            LR &   MLP &    RF &   SVM &             LR &   MLP &    RF &   SVM \\
			classification model       &          &         &         &           &       &       &       &            &       &       &       &       &       &       &       &       &       &       &       &               &       &       &       &                &       &       &       \\
			\midrule
			LR-Confidence  &     -&   - &  - &    {\bf 0.80} & - &  - & - &       0.70 & - & - & - &  0.62 & - &  - & - &  0.70 & - & - &  - &   0.78 & - & - & - &           0.71 & - &  - &  - \\
			SVM-Platt \cite{platt1999probabilistic}  &    -&   -&   - &    - & - &  -&  0.83 &      - & - &  - &  0.75 &  - & - &  - &  0.72 &  - & - &  - &  0.76 &    - &  - &  - &  0.85 &          - &  - & - &  0.75 \\			
			DNN-MCP \cite{hendrycks2016baseline}  &     0.78 &    0.90 &   {\bf 0.98} &   - & {\bf 0.80} & - &  - &    - &  0.76 & - & - & - &  0.57 & - &  - &  - &  0.68 &  - &  - &   - &  0.78 &  - &  - &   - &  0.76 & - &  - \\
			RF-uncertainty \cite{shaker2020aleatoric} &      - &   - &   - &     - & - &  0.75 & - &      - & - &  0.71 & - & - & - &  0.65 & - & - & - &  {\bf 0.75} & - &         - & - &  0.69 & - &          - & - &  0.70 & - \\
			Trust score \cite{jiang2018trust}    &     0.64 &    0.88 &    0.96 &      0.65 &  0.65 &  0.71 &  0.71 &       0.69 &  0.69 &  0.83 &  0.81 & {\bf 0.67} &  0.71 &  0.69 &  0.73 &  0.70 &  0.66 &  0.65 &  0.69 &          0.62 &  0.62 &  0.67 &  0.67 &           0.70 &  0.68 &  0.83 &  0.82 \\
			Riskscore (Proposed)    &    {\bf 0.80} &   {\bf 0.92} &  {\bf 0.98} &     {\bf 0.80} & {\bf 0.80} & {\bf 0.87} & {\bf 0.86} &  {\bf 0.83} & {\bf 0.77} & {\bf 0.86} &  {\bf 0.86} &  0.66 &  {\bf 0.75} & {\bf 0.74} & {\bf 0.75} & {\bf 0.78} &  {\bf 0.72} &  0.72 &  {\bf 0.79} &  {\bf 0.79} &  {\bf 0.79} & {\bf 0.87} &  {\bf 0.87} & {\bf  0.83} &  {\bf 0.77} & {\bf 0.86} & {\bf 0.85} \\
			\bottomrule
		\end{tabular}
	}
	\caption{AUROC for predicting test-time failure risks: 
		Values  in the table are area under ROC curve (AUROC). Higher values are better.} 
	\label{tbl:aucroc}
\end{table*}

\begin{table*}[bth!]
	\centering
	\setlength{\tabcolsep}{3pt}
	\resizebox{\textwidth}{!}{
		\begin{tabular}{l|c|c|c|cccc|cccc|cccc|cccc|cccc|cccc}
			\toprule
			Dataset &  CIFAR 10 & Fashion &   MNIST & \multicolumn{4}{c|}{Census Income} & \multicolumn{4}{c|}{Law School} & \multicolumn{4}{c|}{Wine Quality} & \multicolumn{4}{c|}{Heart Disease} & \multicolumn{4}{c|}{Census Income} & \multicolumn{4}{c}{Law School} \\
			&   & MNIST &    & \multicolumn{4}{c|}{} & \multicolumn{4}{c|}{} & \multicolumn{4}{c|}{white $\rightarrow$ white, red} & \multicolumn{4}{c|}{US $\rightarrow$ US, UK, CH, HU} & \multicolumn{4}{c|}{Male $\rightarrow$ Male, Female} & \multicolumn{4}{c}{White $\rightarrow$ White, Black} \\
			\midrule
			Black-box (BBox)& ResNet50 & CNN & CNN &        LR &   MLP &    RF &   SVM &         LR &   MLP &    RF &   SVM &    LR &   MLP &    RF &   SVM &    LR &   MLP &    RF &   SVM &            LR &   MLP &    RF &   SVM &             LR &   MLP &    RF &   SVM \\
			classification model       &          &         &         &           &       &       &       &            &       &       &       &       &       &       &       &       &       &       &       &               &       &       &       &                &       &       &       \\
			\midrule
			LR-Confidence  &     - &    - &  - &      0.59 & - &  -&  - &       0.39 &  - & - &  - &  0.23 &  - &  - & - &  0.41 & - &  - &  -&          0.56 &  - &  - &  - &           0.41 &  -&  - &  - \\
			SVM-Platt \cite{platt1999probabilistic}  &    - &    - &  - &      - & - & - &  0.66 &  - & - &  - & 0.51 &  - &  - & - & 0.44 &  - &  - &  - & 0.52 &          - & - &  - &  0.69 &         - &  - & - &  0.50 \\
			DNN-MCP \cite{hendrycks2016baseline}  &     0.57 &    0.80 &   {\bf 0.96} &   -    & {\bf 0.60} &  - &  - &       - &  0.52 &  - &  - & - &  0.14 &  - &  - &  - &  0.36 &  - &  - &         - &  0.56 & -&  - &          - &  0.53 & -&  - \\
			RF-uncertainty \cite{shaker2020aleatoric} &      - &   - &    - &      - & - &  0.51 & - &  - & - & 0.41 & - & - & - & 0.30 & - &  - & - & {\bf 0.50} & - &    - &  -& 0.38 & - &           -& - & 0.41 & - \\
			Trust score \cite{jiang2018trust} &     0.29 &    0.77 &    0.91 &      0.31 &  0.30 &  0.41 &  0.42 &       0.38 &  0.38 &  0.67 &  0.63 & {\bf 0.34} &  0.41 &  0.38 &  0.46 &  0.39 &  0.32 &  0.30 &  0.38 &          0.24 &  0.25 &  0.34 &  0.35 &           0.41 &  0.36 &  0.66 &  0.63 \\
			Riskscore (Proposed)       &    {\bf 0.59} &   {\bf 0.83} &    {\bf 0.96} &    {\bf 0.61} &  {\bf 0.60} & {\bf 0.74} &  {\bf 0.73} &      {\bf 0.65} &  {\bf 0.54} &  {\bf 0.73}  &  {\bf 0.71} &  0.32 &  {\bf 0.50} &  {\bf 0.47} &  {\bf 0.50} & {\bf 0.57} & {\bf 0.45} &  0.44 &  {\bf 0.57} &        {\bf  0.57} &  {\bf 0.58} &  {\bf 0.74} &  {\bf 0.74} &        {\bf 0.66} & {\bf 0.53} &  {\bf 0.73} & {\bf 0.70} \\
			\bottomrule
		\end{tabular}
	}
	\caption{Risk mitigation by selective abstention. Values in the table are prediction rejection ratio (PRR). Higher values are better.} 
	\label{tbl:PRR}
	\vspace{-4mm}
\end{table*}

\subsection{\bf \em Predicting Test-time Failure Risks} \label{subsec:detect-errors}
First, we evaluate to what extent the \emph{Risk Advisor} can successfully detect test points misclassified by the underlying ML system.
We measure the quality of failure prediction using standard metrics used in the literature~\cite{hendrycks2016baseline}:
area under ROC curve (AUROC) and area under precision recall curve (AUPR),
where misclassifications are chosen as the positive class. 

\smallskip
\noindent
\textbf{Results:} 
Table \ref{tbl:aucroc} 
shows a comparison between the black-box models' own \emph{confidence scores}
\cite{hendrycks2016baseline, platt1999probabilistic,shaker2020aleatoric}, \emph{Trust Score}~\cite{jiang2018trust} and the \emph{Risk Advisor}'s estimated \emph{risk score}, 
for all combinations of datasets and black-box models. 
Table \ref{tbl:aucroc} reports AUROC
for detecting test-set errors of the underlying black-box classifiers.
Best values
are marked in bold.
We make the following observations.

First, we observe that AUROC
values for all the methods are higher than a {random baseline} (AUROC of 0.5), 
indicating that all the approaches are informative in detecting test errors.
Second, the proposed \emph{risk score} consistently {\em outperforms} black-box models' own {confidence} scores (barring a few exceptions).
This holds true for all families of black-box classifiers including deep neural models and Random Forests, which 
build on
DNN-MCP \cite{hendrycks2016baseline} and RF-uncertainty \cite{shaker2020aleatoric}.
Finally, we observe that our Risk Advisor's \emph{risk scores} consistently \emph{outperform Trust Scores} by a significant margin,
for all the datasets and all families of black-box classifiers.
Similar trends hold for the AUPR metric (not shown for lack of space).

\subsection{\bf \em Application: Risk Mitigation by Selective Abstention}\label{subsec:abstain-from-predictions}

A 
benefit
of predicting failure risks at deployment time
is that we can take meaningful \emph{risk mitigation} actions. For instance, if we expect that a ML system is likely to 
misclassify
certain deployment/test-points, we can ask the ML system to abstain from making predictions and instead forward these data points to a fall-back system or human expert.
In this experiment, we simulate the latter scenario as follows.

\smallskip
\noindent
\textbf{Setup and Metric:} We generate a ranking of all the test points by ordering them according to the scores assigned by each approach, i.e., black-box model's \emph{confidence score} (ascending order), \emph{trust score} (ascending order), and \emph{Risk Advisor}'s \emph{risk score} (descending order), respectively.
We then use these rankings to choose test points to defer to an \emph{oracle}, in which case the ML systems predictions are replaced with the \emph{oracle's} labels. This setup allows us to
compute an \emph{Accuracy-Rejection curve} (AR curve)~\cite{malinin2020uncertainty,bartlett2008classification,el2010foundations}.
AR curves are summarized using \emph{prediction rejection ratio} (PRR), a metric which measures the degree to which the uncertainty scores are informative~\cite{malinin2020uncertainty}.
The \emph{PRR} score lies between 0.0 and 1.0, where 1.0 indicates perfect ordering, and 0.0 indicates ‘random’ ordering.

\smallskip
\noindent
\textbf{Results:}
Table \ref{tbl:PRR} shows a comparison between 
the black-box models' own \emph{confidence scores} \cite{hendrycks2016baseline, platt1999probabilistic,shaker2020aleatoric}, \emph{Trust Scores}~\cite{jiang2018trust} and the proposed \emph{risk scores}.
Values in the table are PRR values for all combinations of datasets and models. Best results are highlighted in bold. We make the following observations.

First, all methods under comparison have a PRR $>0$, indicating that all the approaches are informative and better than a {random baseline} (with random abstention).
Second, \emph{risk scores} consistently yield the \emph{best PRR} across all datasets and classification models (barring a few exceptions). 
There is no clear winner between {Trust Scores} and each of the black-box classifiers' 
native {confidence scores}.

\begin{table*}[tbh!]
	\centering
	\setlength{\tabcolsep}{3.5pt}
	\resizebox{0.72\textwidth}{!}{
		\begin{tabular}{l|cccc|cccc|cccc|cccc}
			\toprule
			Dataset &   \multicolumn{4}{c|}{Wine Quality} & \multicolumn{4}{c|}{Heart Disease} & \multicolumn{4}{c|}{Census Income} & \multicolumn{4}{c}{Law School} \\
			& \multicolumn{4}{c|}{white $\rightarrow$ white, red} & \multicolumn{4}{c|}{US $\rightarrow$ US, UK, CH, HU} & \multicolumn{4}{c|}{Male $\rightarrow$ Male, Female} & \multicolumn{4}{c}{White $\rightarrow$ White, Black} \\
			\midrule
			BBox classification model &    LR &   MLP &    RF &   SVM &    LR &   MLP &    RF &   SVM &            LR &   MLP &    RF &   SVM &             LR &   MLP &    RF &   SVM \\
			\midrule
			LR-Confidence     &  0.33 &  - &  -&  - &  0.66 &  - & - &  - &          0.45 &  - & - &  - &           0.68 &  - &  - &  - \\
			SVM-Platt \cite{platt1999probabilistic}     &  - &  - & - &  0.81 &  - &  - &  - &  0.67 &         - &  -&  - &  0.42 &         - &  - &  - &  0.66 \\
			MCP \cite{hendrycks2016baseline}     &  - &  0.25 &  - & - &  - &  0.63 & - &  - &        - &  0.42 &  - &  - &        -&  0.61 & - &  -\\
			RF-epistemic uncertainty \cite{shaker2020aleatoric}  & - & - & {\bf 0.84} & - & -& - & 0.55 & - &  - & - & {\bf 0.64} & - &  - & - & 0.62 & - \\
			Trust score \cite{jiang2018trust}   &  0.61 &  {0.65} &  0.62 &  0.64 &  0.66 & {\bf 0.65} &  0.62 &  0.64 &          0.42 &  0.42 &  0.42 &  0.42 &           0.66 &  0.68 &  0.67 &  0.62 \\
			Epistemic uncertainty &  {\bf 0.81} &  {\bf 0.87} &  {0.82} & {\bf 0.91} &  {\bf 0.67} &  0.54 &  {\bf 0.74} &  {\bf 0.72} &  {\bf 0.51} & {\bf 0.57} &  {0.54} &  {\bf 0.48} &         {\bf 0.72} & {\bf 0.70} & {\bf 0.72} & {\bf 0.68} \\
			\bottomrule
		\end{tabular}
	}
	\caption{Detecting out-of-distribution (OOD) test examples: Values in the table are AUROC for OOD detection. Higher values are better.}
	\label{tbl:ood-detection}
	\vspace{-4mm}
\end{table*}

\subsection{\bf \em Application: Detecting Out-of-distribution Test Points}\label{subsec:detect-source-of-uncertainty}
In this experiment, we evaluate how well the \emph{Risk Advisor} 
can successfully detect the underlying sources of uncertainty.
However, for real-world datasets and complex black-box models it is difficult to collect ground truth (for evaluation) on which errors are due to inherent data complexity or model limitations.
Hence, in this section we only focus on detecting errors due to \emph{lack of knowledge} 
(e.g., due to data shifts between training and deployment distributions).
To this end, we narrow our focus on the four datasets on out-of-distribution (OOD) test points for which we have ground truth labels shown in Table \ref{tbl:ood-detection}.
Our goal is to evaluate 
how well 
the \emph{Risk Advisor's} estimated {\em epistemic uncertainty} can be used to detect test points coming from a different distribution than the one which the model was trained on.

\smallskip
\noindent
\textbf{Setup and Metric:} Given a combined test dataset consisting of both in-distribution and out-of-distribution test points, the question at hand is to what extent the \emph{Risk Advisor}'s estimated \emph{epistemic uncertainty} can effectively separate in-distribution and out-of-distribution test points.
As we have ground truth for out-of-distribution test points and we have ensured that there are equal numbers of in/out distribution test points, we can use the 
area-under-the-ROC-curve metric (AUROC) for evaluation.
Intuitively, AUROC measures the degree to which each of the confidence scores ranks a randomly chosen OOD data point higher than a randomly chosen non-OOD point. 

\smallskip
\noindent
\textbf{Results:}
Table \ref{tbl:ood-detection} shows a comparison between the black-box model's own \emph{confidence} score, \emph{trust score}, and the \emph{Risk Advisor}'s estimated \emph{epistemic uncertainty}. 
Unlike baseline methods for DNN, SVM, and LR, the baseline for computing uncertainty of RF by \cite{shaker2020aleatoric} can decompose the overall uncertainty into aleatoric and epistemic components. Thus, for RF, we rely on the {\em epistemic} uncertainty estimates.
We make the following observations.

First, observe that \emph{epistemic uncertainty} consistently outperforms both the black-box model's own \emph{confidence scores} and \emph{trust scores} across all datasets and classification methods, with a significant margin.
Further {\em Risk Advisor's} epistemic uncertainty is competitive with {RF-epistemic uncertainty}, 
which 
is model-specific.
This supports our argument that a post-hoc {\em meta-learner} trained to compute uncertainties, is a viable 
alternative to replacing the underlying black-box ML classifier, which may not be feasible in production practice.
Second, observe that for the Wine and Census Income datasets,
the  DNN-MCP\cite{hendrycks2016baseline} and LR \emph{confidence score} has AUROC $<$0.5, i.e., a performance worse than the {\em random baseline}, implying that black-box model \emph{incorrectly} assigns higher confidence scores for OOD points than for in-distribution points.
A similar trend can be observed for \emph{trust scores} for the Census Income dataset, 
thus indicating that \emph{confidence} scores and \emph{trust scores} are not that reliable under distribution shifts. 
In contrast, the the AUROC values for the \emph{epistemic uncertainty} are always $>0.5$,
implying that the \emph{Risk Advisor} always assigns higher \emph{epistemic} uncertainty for OOD test points than for in-distribution test points. 
This is an important property, as it indicates that a ranked ordering of test points by \emph{epistemic uncertainty} can be used in a deployed application to detect out-of-distribution test points (given an application-specific threshold). For these critical data points,
the system could resort to a human expert (or other fall-back option),
and thus enhance 
trustworthiness of the ML system.

\begin{figure}[tbh!]
	\begin{subfigure}{0.49\columnwidth}
		\centering	
		\includegraphics[scale=0.39]{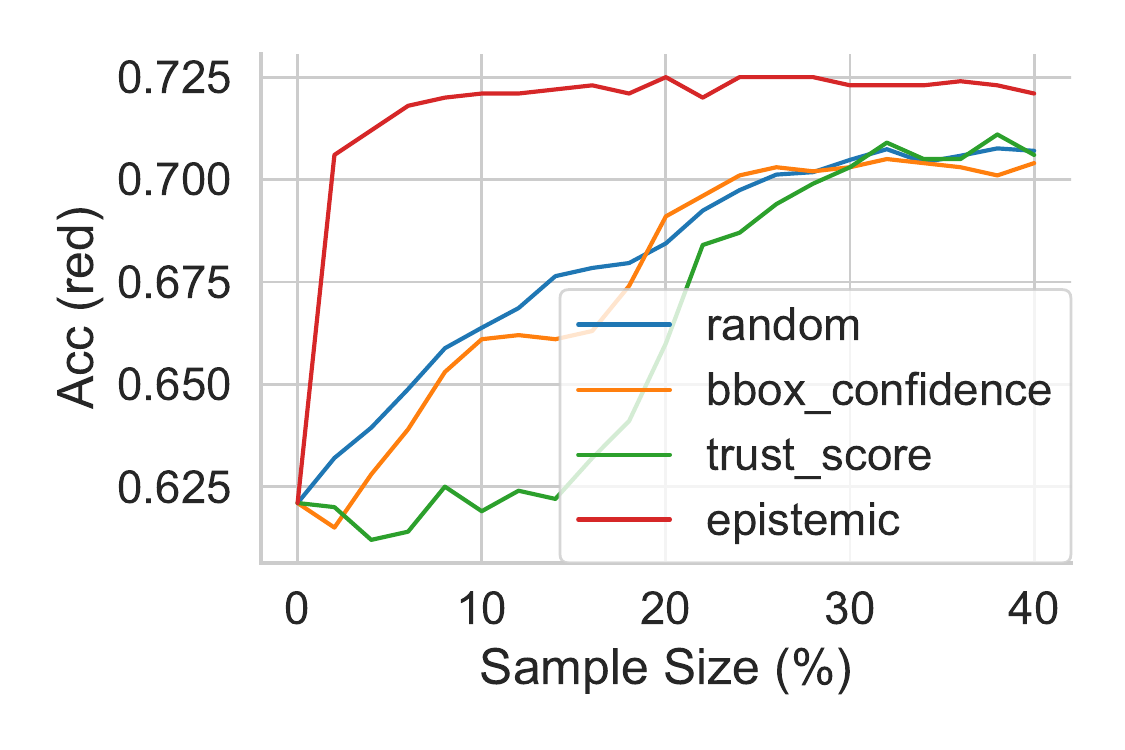}
		\vspace{-6mm}%
		\caption{Wine Quality}
	\end{subfigure}
	\begin{subfigure}{0.49\columnwidth}
		\centering	
		\includegraphics[scale=0.39]{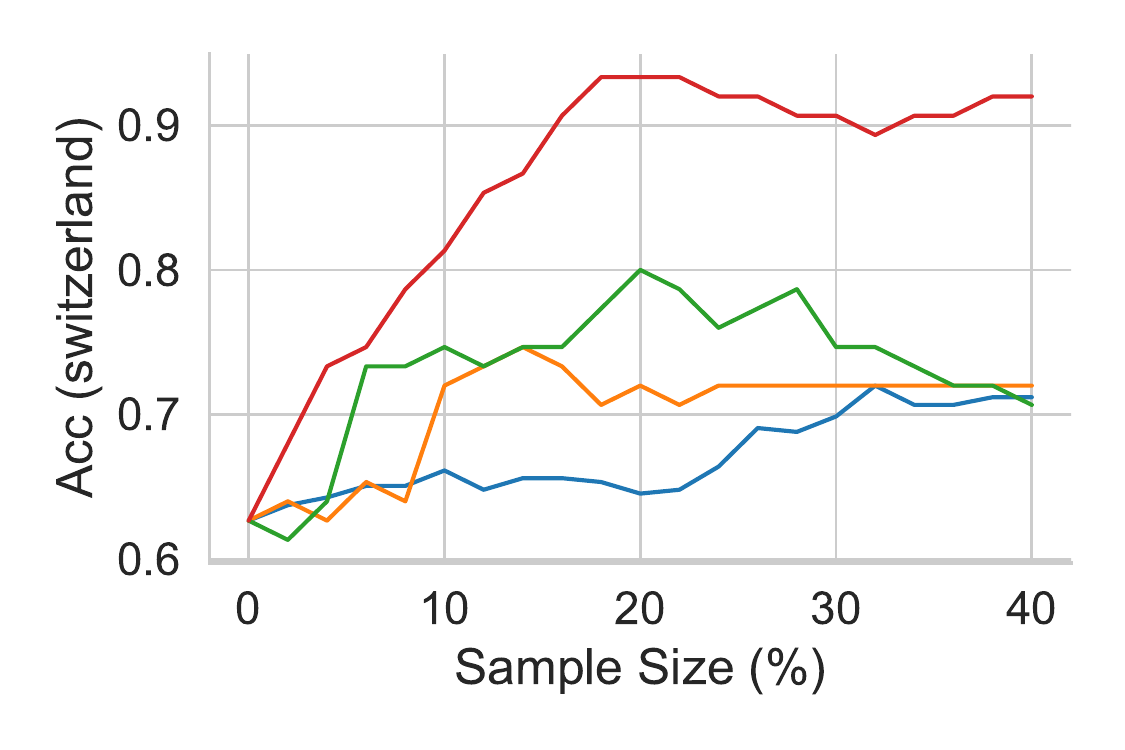}
		\vspace{-6mm}%
		\caption{Heart Disease}
	\end{subfigure}
	\newline
	\begin{subfigure}{0.49\columnwidth}
		\centering	
		\includegraphics[scale=0.39]{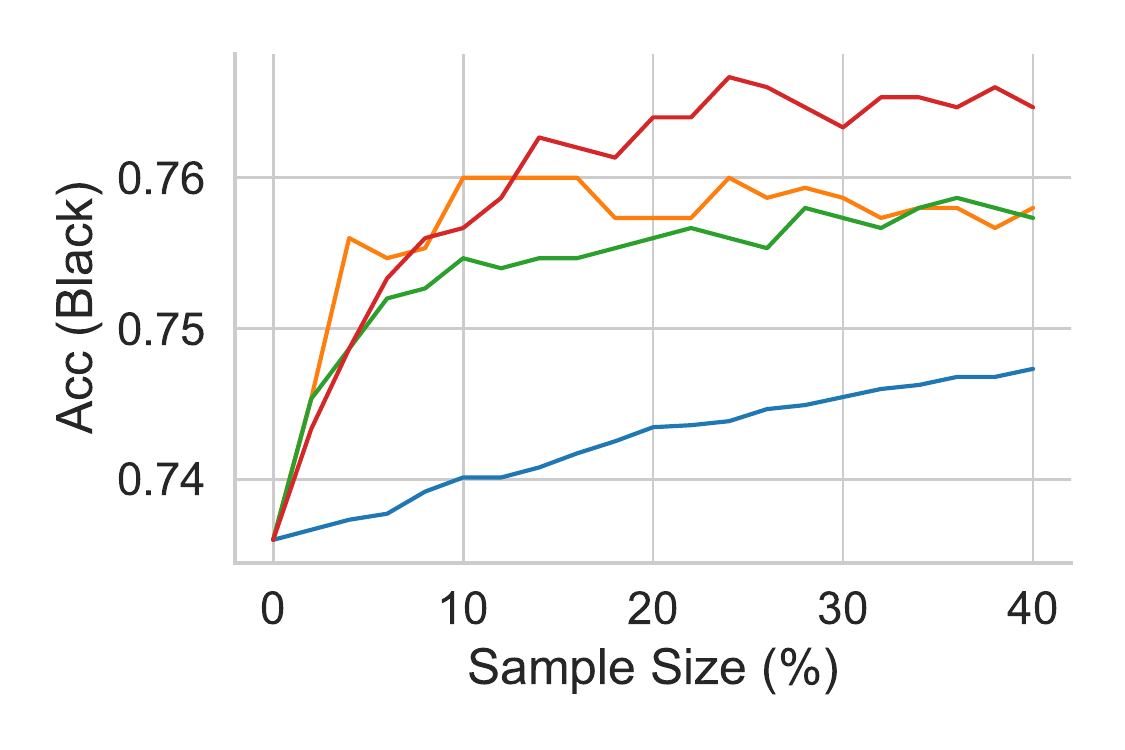}
		\vspace{-6mm}%
		\caption{Law School (White $\rightarrow$ Black)}
	\end{subfigure}			
	\begin{subfigure}{0.49\columnwidth}
		\centering	
		\includegraphics[scale=0.39]{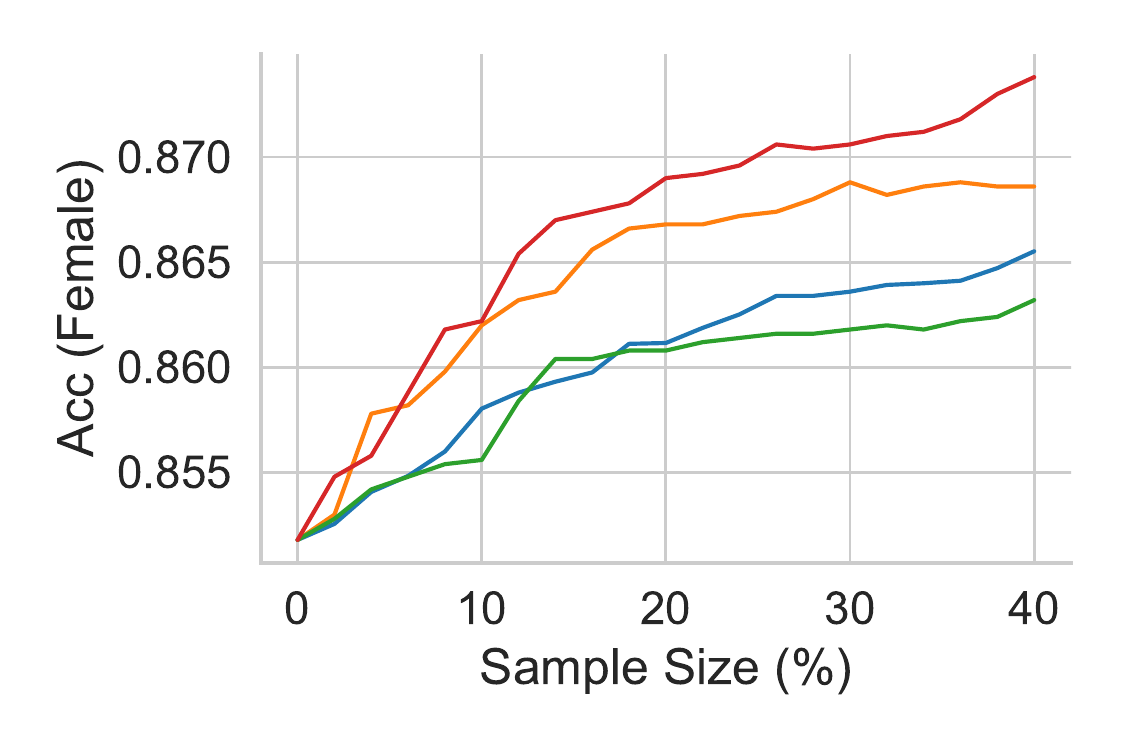}
		\vspace{-6mm}%
		\caption{Census Income (Male $\rightarrow$ Female)}
	\end{subfigure}	
	\caption{Addressing distribution shift: Comparison of various sampling strategies to selectively sample data points and retrain the black-box classification model. Curves that grow higher and faster from left to right are better.}
	\label{fig:sample-retrain}
	\vspace{-4mm}
\end{figure}

\subsection{\bf \em Application: Risk Mitigation by Sampling \& Retraining}\label{subsec:sample-retrain}
Being able to identify black-box classifier's epistemic uncertainty
enables another type of mitigation action: to mitigate risks due to evolving data by judiciously collecting more training examples and re-training the ML system.

We acknowledge the large body of 
literature on active sampling and domain adaptation in this context.
In our experiment the goal is not to compare with these existing techniques, 
but rather to demonstrate an application of the \emph{Risk Advisor}'s epistemic uncertainty,
which can be achieved without making any changes to the underlying black-box classification system. %

\smallskip
\noindent
\textbf{Setup:} 
In this experiment, 
we fix the black-box classifier to logistic regression, and we assume that we have access to an untouched held-out set of labeled samples (different from training and test set).
Our goal is to evaluate if the performance of the underlying black-box classifier can be improved for out-of-distribution test points by additional sampling and re-training the ML system on 
(a subset of) these held-out points.  To evaluate the performance, we use the black-box classifier's improvement in accuracy for out-of-distribution test points.

We compare different sampling strategies by selecting data points from the with-held set
in different orders based on three criteria:
the \emph{LR-Confidence}, \emph{Trust score}, and 
the \emph{Risk Advisor}'s \emph{epistemic uncertainty}.
For each approach, we first compute point-wise scores for all the points in the held-out set (different from training and test set, kept aside for sampling experiment). 
We then order the points in the held-out set according to these scores, i.e., \emph{LR-confidence} (ascending order), \emph{Trust score} (ascending order), and \emph{Risk Advisor}'s \emph{epistemic uncertainty} (descending order), respectively.
Next, at each round of an iterative sampling, we select $k\%$ points from the held-out set (with replacement), 
and re-train the ML system.

\smallskip
\noindent
\textbf{Results:} 
Fig. \ref{fig:sample-retrain} shows results averaged over 5 independent runs. 
The x-axis shows the percentage of additional points sampled from the held-out set for re-training, and the y-axis shows the corresponding improved accuracy for the OOD group (e.g., accuracy on red wine for the Wine dataset). Ideally, we would expect the accuracy to rise higher with as few additional training points as possible. We make the following observations.

First, as we sample and retrain on additional points from the held-out data, the accuracy for OOD test-points increases for all the approaches on all datasets. However, the percentage of additional samples required to achieve similar performance differs across approaches. 
Not surprisingly, \emph{random} sampling is the slowest improving approach for 3 out of 4 datasets, followed by \emph{trust scores} and \emph{confidence scores}. 
The Risk Advisor's sampling by \emph{epistemic} uncertainty consistently outperforms
on all datasets, by a large margin.
For instance, on the Heart Disease dataset \emph{epistemic} uncertainty achieves 30 percentage points (pp) improvement in accuracy (from 0.6 to 0.9) for an additional $20\%$ samples from the held-out set. In contrast, all the other approaches stagnate around $0.7$ even for an additional $40\%$ samples. Similarly, on the Wine Quality dataset
we see an improvement of 10 pp for an additional $10\%$ samples, while 
other approaches do not reach this improvement even for additional $40\%$ of samples.
We observe similar trends across approaches for Law School and Census Income datasets, albeit with smaller
gains.

\section{Conclusion}

This paper presented the \emph{Risk Advisor} model for detecting and analyzing
sources of uncertainty and failure risks when a trained classifier is deployed for production usage.
In contrast to prior works, the Risk Advisor treats the base classifier as a black-box model,
and this model-agnostic approach makes it a highly versatile and easy-to-deploy tool.
In contrast to the 
prior state-of-the-art (including the main baseline Trust Scores \cite{jiang2018trust}),
the Risk Advisor goes beyond providing a single measure of uncertainty,
by computing refined scores that indicate failure risks due to 
data variability and noise, systematic data shifts between training and deployment,
and model limitations.
Extensive experiments 
on real-world and synthetic datasets covering common ML failure scenarios show that the Risk Advisor reliably predicts deployment-time failure risks in all the scenarios, and outperforms strong baselines.
Thereby, we believe the Risk advisor, with its ability to proactively audit and identify potential regions of failure risks would be a useful asset
for the trustworthy machine learning toolbox.

\section{Acknowledgment} This research was supported by the ERC Synergy Grant ``imPACT'' (No. 610150) and ERC Advanced Grant ``Foundations for Fair Social Computing'' (No. 789373).
\balance
\bibliographystyle{ACM-Reference-Format}
\bibliography{references}


\begin{thebibliography}{39}


\ifx \showCODEN    \undefined \def \showCODEN     #1{\unskip}     \fi
\ifx \showDOI      \undefined \def \showDOI       #1{#1}\fi
\ifx \showISBNx    \undefined \def \showISBNx     #1{\unskip}     \fi
\ifx \showISBNxiii \undefined \def \showISBNxiii  #1{\unskip}     \fi
\ifx \showISSN     \undefined \def \showISSN      #1{\unskip}     \fi
\ifx \showLCCN     \undefined \def \showLCCN      #1{\unskip}     \fi
\ifx \shownote     \undefined \def \shownote      #1{#1}          \fi
\ifx \showarticletitle \undefined \def \showarticletitle #1{#1}   \fi
\ifx \showURL      \undefined \def \showURL       {\relax}        \fi
\providecommand\bibfield[2]{#2}
\providecommand\bibinfo[2]{#2}
\providecommand\natexlab[1]{#1}
\providecommand\showeprint[2][]{arXiv:#2}

\bibitem[\protect\citeauthoryear{Barber and Bishop}{Barber and Bishop}{1998}]%
        {barber1998ensemble}
\bibfield{author}{\bibinfo{person}{D. Barber} {and} \bibinfo{person}{C.~M
  Bishop}.} \bibinfo{year}{1998}\natexlab{}.
\newblock \showarticletitle{Ensemble learning in Bayesian neural networks}.
\newblock \bibinfo{journal}{\emph{J. Comput. Syst. Sci.}}
  \bibinfo{volume}{168} (\bibinfo{year}{1998}).
\newblock


\bibitem[\protect\citeauthoryear{Bartlett and Wegkamp}{Bartlett and
  Wegkamp}{2008}]%
        {bartlett2008classification}
\bibfield{author}{\bibinfo{person}{P.~L Bartlett} {and} \bibinfo{person}{M.~H
  Wegkamp}.} \bibinfo{year}{2008}\natexlab{}.
\newblock \showarticletitle{Classification with a reject option using a hinge
  loss}.
\newblock \bibinfo{journal}{\emph{JMLR}} \bibinfo{volume}{9},
  \bibinfo{number}{Aug} (\bibinfo{year}{2008}).
\newblock


\bibitem[\protect\citeauthoryear{Ben-Gal}{Ben-Gal}{2005}]%
        {ben2005outlier}
\bibfield{author}{\bibinfo{person}{Irad Ben-Gal}.}
  \bibinfo{year}{2005}\natexlab{}.
\newblock \showarticletitle{Outlier detection}.
\newblock In \bibinfo{booktitle}{\emph{Data mining and knowledge discovery
  handbook}}. \bibinfo{publisher}{Springer}.
\newblock


\bibitem[\protect\citeauthoryear{Cover}{Cover}{1999}]%
        {cover1999elements}
\bibfield{author}{\bibinfo{person}{Thomas~M Cover}.}
  \bibinfo{year}{1999}\natexlab{}.
\newblock \bibinfo{booktitle}{\emph{Elements of information theory}}.
\newblock \bibinfo{publisher}{John Wiley \& Sons}.
\newblock


\bibitem[\protect\citeauthoryear{Denker and LeCun}{Denker and LeCun}{1990}]%
        {denker1990transforming}
\bibfield{author}{\bibinfo{person}{J.~S Denker} {and} \bibinfo{person}{Y.
  LeCun}.} \bibinfo{year}{1990}\natexlab{}.
\newblock \showarticletitle{Transforming neural-net output levels to
  probability distributions}. In \bibinfo{booktitle}{\emph{NeurIPS}}.
\newblock


\bibitem[\protect\citeauthoryear{Depeweg et~al\mbox{.}}{Depeweg
  et~al\mbox{.}}{2018}]%
        {depeweg2018decomposition}
\bibfield{author}{\bibinfo{person}{S. Depeweg} {et~al\mbox{.}}}
  \bibinfo{year}{2018}\natexlab{}.
\newblock \showarticletitle{Decomposition of uncertainty in Bayesian deep
  learning for efficient and risk-sensitive learning}. In
  \bibinfo{booktitle}{\emph{ICML}}.
\newblock


\bibitem[\protect\citeauthoryear{Der~Kiureghian and Ditlevsen}{Der~Kiureghian
  and Ditlevsen}{2009}]%
        {der2009aleatory}
\bibfield{author}{\bibinfo{person}{A. Der~Kiureghian} {and} \bibinfo{person}{O.
  Ditlevsen}.} \bibinfo{year}{2009}\natexlab{}.
\newblock \showarticletitle{Aleatory or epistemic? Does it matter?}
\newblock \bibinfo{journal}{\emph{Structural safety}} \bibinfo{volume}{31},
  \bibinfo{number}{2} (\bibinfo{year}{2009}).
\newblock


\bibitem[\protect\citeauthoryear{Dua and Graff}{Dua and Graff}{2017}]%
        {uci}
\bibfield{author}{\bibinfo{person}{D. Dua} {and} \bibinfo{person}{C. Graff}.}
  \bibinfo{year}{2017}\natexlab{}.
\newblock \bibinfo{title}{{UCI} Machine Learning Repository}.
\newblock
\newblock
\urldef\tempurl%
\url{http://archive.ics.uci.edu/ml}
\showURL{%
\tempurl}


\bibitem[\protect\citeauthoryear{El-Yaniv et~al\mbox{.}}{El-Yaniv
  et~al\mbox{.}}{2010}]%
        {el2010foundations}
\bibfield{author}{\bibinfo{person}{R. El-Yaniv} {et~al\mbox{.}}}
  \bibinfo{year}{2010}\natexlab{}.
\newblock \showarticletitle{On the Foundations of Noise-free Selective
  Classification.}
\newblock \bibinfo{journal}{\emph{JMLR}} \bibinfo{volume}{11},
  \bibinfo{number}{5} (\bibinfo{year}{2010}).
\newblock


\bibitem[\protect\citeauthoryear{Fano}{Fano}{1961}]%
        {fano1961transmission}
\bibfield{author}{\bibinfo{person}{R.~M Fano}.}
  \bibinfo{year}{1961}\natexlab{}.
\newblock \showarticletitle{Transmission of information: A statistical theory
  of communications}.
\newblock \bibinfo{journal}{\emph{American Journal of Physics}}
  \bibinfo{volume}{29}, \bibinfo{number}{11} (\bibinfo{year}{1961}).
\newblock


\bibitem[\protect\citeauthoryear{Friedman}{Friedman}{2002}]%
        {friedman2002stochastic}
\bibfield{author}{\bibinfo{person}{J.~H Friedman}.}
  \bibinfo{year}{2002}\natexlab{}.
\newblock \showarticletitle{Stochastic gradient boosting}.
\newblock \bibinfo{journal}{\emph{Computational statistics \& data analysis}}
  \bibinfo{volume}{38}, \bibinfo{number}{4} (\bibinfo{year}{2002}).
\newblock


\bibitem[\protect\citeauthoryear{Gal and Ghahramani}{Gal and
  Ghahramani}{2016}]%
        {gal2016dropout}
\bibfield{author}{\bibinfo{person}{Y. Gal} {and} \bibinfo{person}{Z.
  Ghahramani}.} \bibinfo{year}{2016}\natexlab{}.
\newblock \showarticletitle{Dropout as a bayesian approximation: Representing
  model uncertainty in deep learning}. In \bibinfo{booktitle}{\emph{ICML}}.
\newblock


\bibitem[\protect\citeauthoryear{Goodfellow, Shlens, and Szegedy}{Goodfellow
  et~al\mbox{.}}{2015}]%
        {goodfellow2014explaining}
\bibfield{author}{\bibinfo{person}{I.~J Goodfellow}, \bibinfo{person}{J.
  Shlens}, {and} \bibinfo{person}{C. Szegedy}.}
  \bibinfo{year}{2015}\natexlab{}.
\newblock \showarticletitle{Explaining and harnessing adversarial examples}. In
  \bibinfo{booktitle}{\emph{ICLR}}.
\newblock


\bibitem[\protect\citeauthoryear{Guo et~al\mbox{.}}{Guo et~al\mbox{.}}{2017}]%
        {guo2017calibration}
\bibfield{author}{\bibinfo{person}{C. Guo} {et~al\mbox{.}}}
  \bibinfo{year}{2017}\natexlab{}.
\newblock \showarticletitle{On Calibration of Modern Neural Networks}. In
  \bibinfo{booktitle}{\emph{ICML}}.
\newblock


\bibitem[\protect\citeauthoryear{He, Zhang, Ren, and Sun}{He
  et~al\mbox{.}}{2016}]%
        {resnet50}
\bibfield{author}{\bibinfo{person}{K. He}, \bibinfo{person}{X. Zhang},
  \bibinfo{person}{S. Ren}, {and} \bibinfo{person}{J. Sun}.}
  \bibinfo{year}{2016}\natexlab{}.
\newblock \showarticletitle{Deep residual learning for image recognition}. In
  \bibinfo{booktitle}{\emph{CVPR}}.
\newblock


\bibitem[\protect\citeauthoryear{Hendrycks and Gimpel}{Hendrycks and
  Gimpel}{2017}]%
        {hendrycks2016baseline}
\bibfield{author}{\bibinfo{person}{D. Hendrycks} {and} \bibinfo{person}{K.
  Gimpel}.} \bibinfo{year}{2017}\natexlab{}.
\newblock \showarticletitle{A baseline for detecting misclassified and
  out-of-distribution examples in neural networks}.
\newblock \bibinfo{journal}{\emph{ICLR}}.
\newblock


\bibitem[\protect\citeauthoryear{Honkela and Valpola}{Honkela and
  Valpola}{2004}]%
        {honkela2004variational}
\bibfield{author}{\bibinfo{person}{A. Honkela} {and} \bibinfo{person}{H.
  Valpola}.} \bibinfo{year}{2004}\natexlab{}.
\newblock \showarticletitle{Variational learning and bits-back coding: an
  information-theoretic view to Bayesian learning}.
\newblock \bibinfo{journal}{\emph{IEEE Trans. Neural Netw.}}
  \bibinfo{volume}{15}, \bibinfo{number}{4} (\bibinfo{year}{2004}).
\newblock


\bibitem[\protect\citeauthoryear{Hora}{Hora}{1996}]%
        {hora1996aleatory}
\bibfield{author}{\bibinfo{person}{Stephen~C Hora}.}
  \bibinfo{year}{1996}\natexlab{}.
\newblock \showarticletitle{Aleatory and epistemic uncertainty in probability
  elicitation with an example from hazardous waste management}.
\newblock \bibinfo{journal}{\emph{Reliability Engineering \& System Safety}}
  \bibinfo{volume}{54}, \bibinfo{number}{2-3} (\bibinfo{year}{1996}).
\newblock


\bibitem[\protect\citeauthoryear{H{\"u}llermeier and Waegeman}{H{\"u}llermeier
  and Waegeman}{2021}]%
        {hullermeier2021aleatoric}
\bibfield{author}{\bibinfo{person}{E. H{\"u}llermeier} {and}
  \bibinfo{person}{W. Waegeman}.} \bibinfo{year}{2021}\natexlab{}.
\newblock \showarticletitle{Aleatoric and epistemic uncertainty in machine
  learning: An introduction to concepts and methods}.
\newblock \bibinfo{journal}{\emph{Machine Learning}} \bibinfo{volume}{110},
  \bibinfo{number}{3} (\bibinfo{year}{2021}).
\newblock


\bibitem[\protect\citeauthoryear{Jiang, Kim, Guan, and Gupta}{Jiang
  et~al\mbox{.}}{2018}]%
        {jiang2018trust}
\bibfield{author}{\bibinfo{person}{H. Jiang}, \bibinfo{person}{B. Kim},
  \bibinfo{person}{M.~Y Guan}, {and} \bibinfo{person}{M.~R Gupta}.}
  \bibinfo{year}{2018}\natexlab{}.
\newblock \showarticletitle{To Trust Or Not To Trust A Classifier.}. In
  \bibinfo{booktitle}{\emph{NeurIPS}}.
\newblock


\bibitem[\protect\citeauthoryear{Kendall and Gal}{Kendall and Gal}{2017}]%
        {kendall2017uncertainties}
\bibfield{author}{\bibinfo{person}{A. Kendall} {and} \bibinfo{person}{Y. Gal}.}
  \bibinfo{year}{2017}\natexlab{}.
\newblock \showarticletitle{What uncertainties do we need in Bayesian deep
  learning for computer vision?}. In \bibinfo{booktitle}{\emph{NeurIPS}}.
\newblock


\bibitem[\protect\citeauthoryear{Krizhevsky}{Krizhevsky}{2009}]%
        {cifar10}
\bibfield{author}{\bibinfo{person}{A. Krizhevsky}.}
  \bibinfo{year}{2009}\natexlab{}.
\newblock \bibinfo{booktitle}{\emph{Learning multiple layers of features from
  tiny images}}.
\newblock \bibinfo{type}{{T}echnical {R}eport}.
\newblock


\bibitem[\protect\citeauthoryear{Lakshminarayanan, Pritzel, and
  Blundell}{Lakshminarayanan et~al\mbox{.}}{2017}]%
        {lakshminarayanan2017simple}
\bibfield{author}{\bibinfo{person}{B. Lakshminarayanan}, \bibinfo{person}{A.
  Pritzel}, {and} \bibinfo{person}{C. Blundell}.}
  \bibinfo{year}{2017}\natexlab{}.
\newblock \showarticletitle{Simple and scalable predictive uncertainty
  estimation using deep ensembles}. In \bibinfo{booktitle}{\emph{NeurIPS}}.
\newblock


\bibitem[\protect\citeauthoryear{LeCun, Cortes, and Burges}{LeCun
  et~al\mbox{.}}{2010}]%
        {lecun2010mnist}
\bibfield{author}{\bibinfo{person}{Yann LeCun}, \bibinfo{person}{Corinna
  Cortes}, {and} \bibinfo{person}{CJ Burges}.} \bibinfo{year}{2010}\natexlab{}.
\newblock \showarticletitle{MNIST handwritten digit database}.
\newblock \bibinfo{journal}{\emph{Available: http://yann.lecun.com/exdb/mnist}}
   \bibinfo{volume}{2} (\bibinfo{year}{2010}).
\newblock


\bibitem[\protect\citeauthoryear{Malinin}{Malinin}{2019}]%
        {malinin2020uncertainty}
\bibfield{author}{\bibinfo{person}{A. Malinin}.}
  \bibinfo{year}{2019}\natexlab{}.
\newblock \showarticletitle{Uncertainty Estimation in Deep Learning with
  application to Spoken LanguageAssessment}.
\newblock \bibinfo{journal}{\emph{PhD thesis}} (\bibinfo{year}{2019}).
\newblock


\bibitem[\protect\citeauthoryear{Malinin, Prokhorenkova, and Ustimenko}{Malinin
  et~al\mbox{.}}{2020}]%
        {ustimenko2020uncertainty}
\bibfield{author}{\bibinfo{person}{A. Malinin}, \bibinfo{person}{L.
  Prokhorenkova}, {and} \bibinfo{person}{A. Ustimenko}.}
  \bibinfo{year}{2020}\natexlab{}.
\newblock \showarticletitle{Uncertainty in gradient boosting via ensembles}.
\newblock \bibinfo{journal}{\emph{arXiv preprint arXiv:2006.10562}}.
\newblock


\bibitem[\protect\citeauthoryear{Nguyen, Yosinski, and Clune}{Nguyen
  et~al\mbox{.}}{2015}]%
        {nguyen2015deep}
\bibfield{author}{\bibinfo{person}{A. Nguyen}, \bibinfo{person}{J. Yosinski},
  {and} \bibinfo{person}{J. Clune}.} \bibinfo{year}{2015}\natexlab{}.
\newblock \showarticletitle{Deep neural networks are easily fooled: High
  confidence predictions for unrecognizable images}. In
  \bibinfo{booktitle}{\emph{CVPR}}.
\newblock


\bibitem[\protect\citeauthoryear{Ovadia et~al\mbox{.}}{Ovadia
  et~al\mbox{.}}{2019}]%
        {ovadia2019can}
\bibfield{author}{\bibinfo{person}{Y. Ovadia} {et~al\mbox{.}}}
  \bibinfo{year}{2019}\natexlab{}.
\newblock \showarticletitle{Can you trust your model's uncertainty? Evaluating
  predictive uncertainty under dataset shift}.
\newblock \bibinfo{journal}{\emph{NeurIPS}}.
\newblock


\bibitem[\protect\citeauthoryear{Pedregosa et~al\mbox{.}}{Pedregosa
  et~al\mbox{.}}{2011}]%
        {scikit-learn}
\bibfield{author}{\bibinfo{person}{F. Pedregosa} {et~al\mbox{.}}}
  \bibinfo{year}{2011}\natexlab{}.
\newblock \showarticletitle{Scikit-learn: ML in {P}ython}.
\newblock \bibinfo{journal}{\emph{JMLR}}  \bibinfo{volume}{12}
  (\bibinfo{year}{2011}).
\newblock


\bibitem[\protect\citeauthoryear{Platt et~al\mbox{.}}{Platt
  et~al\mbox{.}}{1999}]%
        {platt1999probabilistic}
\bibfield{author}{\bibinfo{person}{J. Platt} {et~al\mbox{.}}}
  \bibinfo{year}{1999}\natexlab{}.
\newblock \showarticletitle{Probabilistic outputs for support vector machines
  and comparisons to regularized likelihood methods}.
\newblock \bibinfo{journal}{\emph{Advances in large margin classifiers}}
  \bibinfo{volume}{10}, \bibinfo{number}{3} (\bibinfo{year}{1999}).
\newblock


\bibitem[\protect\citeauthoryear{Saria and Subbaswamy}{Saria and
  Subbaswamy}{2019}]%
        {saria2019tutorial}
\bibfield{author}{\bibinfo{person}{S Saria} {and} \bibinfo{person}{A.
  Subbaswamy}.} \bibinfo{year}{2019}\natexlab{}.
\newblock \showarticletitle{Tutorial: safe and reliable machine learning}.
\newblock \bibinfo{journal}{\emph{arXiv preprint arXiv:1904.07204}}
  (\bibinfo{year}{2019}).
\newblock


\bibitem[\protect\citeauthoryear{Schelter, Rukat, and Biessmann}{Schelter
  et~al\mbox{.}}{2020}]%
        {schelter2020learning}
\bibfield{author}{\bibinfo{person}{S. Schelter}, \bibinfo{person}{T. Rukat},
  {and} \bibinfo{person}{F. Biessmann}.} \bibinfo{year}{2020}\natexlab{}.
\newblock \showarticletitle{Learning to Validate the Predictions of Black Box
  Classifiers on Unseen Data}. In \bibinfo{booktitle}{\emph{SIGMOD}}.
\newblock


\bibitem[\protect\citeauthoryear{Schneider, Rusak, Eck, and
  Bringmann}{Schneider et~al\mbox{.}}{2020}]%
        {schneider2020improving}
\bibfield{author}{\bibinfo{person}{S. Schneider}, \bibinfo{person}{E. Rusak},
  \bibinfo{person}{L. Eck}, {and} \bibinfo{person}{O.~et~al. Bringmann}.}
  \bibinfo{year}{2020}\natexlab{}.
\newblock \showarticletitle{Improving robustness against common corruptions by
  covariate shift adaptation}.
\newblock \bibinfo{journal}{\emph{In NeurIPS}}.
\newblock


\bibitem[\protect\citeauthoryear{Schulam and Saria}{Schulam and Saria}{2019}]%
        {schulam2019can}
\bibfield{author}{\bibinfo{person}{P. Schulam} {and} \bibinfo{person}{S.
  Saria}.} \bibinfo{year}{2019}\natexlab{}.
\newblock \showarticletitle{Can you trust this prediction? Auditing pointwise
  reliability after learning}. In \bibinfo{booktitle}{\emph{AISTATS}}.
\newblock


\bibitem[\protect\citeauthoryear{Senge, B{\"o}sner, Dembczy{\'n}ski,
  Haasenritter, and Hirsch}{Senge et~al\mbox{.}}{2014}]%
        {senge2014reliable}
\bibfield{author}{\bibinfo{person}{R. Senge}, \bibinfo{person}{S. B{\"o}sner},
  \bibinfo{person}{K. Dembczy{\'n}ski}, \bibinfo{person}{J. Haasenritter},
  {and} \bibinfo{person}{O.~et~al. Hirsch}.} \bibinfo{year}{2014}\natexlab{}.
\newblock \showarticletitle{Reliable classification: Learning classifiers that
  distinguish aleatoric and epistemic uncertainty}.
\newblock \bibinfo{journal}{\emph{Information Sciences}}  \bibinfo{volume}{255}
  (\bibinfo{year}{2014}).
\newblock


\bibitem[\protect\citeauthoryear{Shaker and H{\"u}llermeier}{Shaker and
  H{\"u}llermeier}{2020}]%
        {shaker2020aleatoric}
\bibfield{author}{\bibinfo{person}{M.~H. Shaker} {and} \bibinfo{person}{E.
  H{\"u}llermeier}.} \bibinfo{year}{2020}\natexlab{}.
\newblock \showarticletitle{Aleatoric and epistemic uncertainty with random
  forests}. In \bibinfo{booktitle}{\emph{IDA}}.
\newblock


\bibitem[\protect\citeauthoryear{Steinwart, Hush, and Scovel}{Steinwart
  et~al\mbox{.}}{2005}]%
        {steinwart2005classification}
\bibfield{author}{\bibinfo{person}{I. Steinwart}, \bibinfo{person}{D. Hush},
  {and} \bibinfo{person}{C. Scovel}.} \bibinfo{year}{2005}\natexlab{}.
\newblock \showarticletitle{A Classification Framework for Anomaly Detection.}
\newblock \bibinfo{journal}{\emph{JMLR}} \bibinfo{volume}{6},
  \bibinfo{number}{2} (\bibinfo{year}{2005}).
\newblock


\bibitem[\protect\citeauthoryear{Wightman}{Wightman}{1998}]%
        {wightman1998lsac}
\bibfield{author}{\bibinfo{person}{L.~F Wightman}.}
  \bibinfo{year}{1998}\natexlab{}.
\newblock \showarticletitle{LSAC National Longitudinal Bar Passage Study. LSAC
  Research Report Series.}
\newblock  (\bibinfo{year}{1998}).
\newblock


\bibitem[\protect\citeauthoryear{Xiao, Rasul, and Vollgraf}{Xiao
  et~al\mbox{.}}{2017}]%
        {fashion-mnist}
\bibfield{author}{\bibinfo{person}{H. Xiao}, \bibinfo{person}{K. Rasul}, {and}
  \bibinfo{person}{R. Vollgraf}.} \bibinfo{year}{2017}\natexlab{}.
\newblock \showarticletitle{Fashion-mnist: a novel image dataset for
  benchmarking machine learning algorithms}.
\newblock \bibinfo{journal}{\emph{arXiv preprint arXiv:1708.07747}}
  (\bibinfo{year}{2017}).
\newblock


\end{thebibliography}
\end{document}